\documentclass[10pt,journal,compsoc]{IEEEtran}
\usepackage{booktabs} 
\usepackage{epstopdf} 
\usepackage{verbatim} 
\usepackage{graphicx}
\usepackage{subfigure} 
\usepackage[linesnumbered,ruled]{algorithm2e} 
\usepackage{multirow} 
\usepackage{adjustbox} 
\usepackage{xcolor}
\usepackage{array}
\usepackage{enumitem}
\newcommand{\tabincell}[2]{\begin{tabular}{@{}#1@{}}#2\end{tabular}} 
\usepackage{threeparttable} 
\usepackage{mathtools,amssymb}
\usepackage[switch]{lineno}

\ifCLASSOPTIONcompsoc
\usepackage[nocompress]{cite}
\else
\usepackage{cite}
\fi

\ifCLASSINFOpdf
\else
\fi

\begin{document}
	\title{\textcolor{black}{Graph} Transfer Learning via Adversarial Domain Adaptation with Graph Convolution}
	
	\author{Quanyu~Dai,
		Xiao-Ming~Wu,
		Jiaren~Xiao,
		Xiao~Shen,
		and Dan~Wang

\IEEEcompsocitemizethanks{\IEEEcompsocthanksitem Xiao Shen (corresponding author) is with the School of Computer Science and Technology, Hainan University, Haikou, China. \protect\\ Email: shenxiaocam@163.com

\IEEEcompsocthanksitem Quanyu Dai is with Huawei Noah’s Ark Lab, Shenzhen, China.\protect\\ Email: quanyu.dai@connect.polyu.hk

\IEEEcompsocthanksitem Xiao-Ming Wu and Dan Wang are with the Department of Computing, The Hong Kong Polytechnic University, Kowloon 999077, Hong Kong.\protect\\ Email: \{csxmwu, csdwang\}@comp.polyu.edu.hk

\IEEEcompsocthanksitem Jiaren Xiao is with with the Department of Mechanical Engineering, The University of Hong Kong, Pokfulam 999077, Hong Kong.\protect\\ Email: xiaojr@connect.hku.hk}
	}

	\markboth{IEEE TRANSACTIONS ON KNOWLEDGE AND DATA ENGINEERING, Submission 2019}%
	{Shell \MakeLowercase{\textit{et al.}}: Bare Demo of IEEEtran.cls for Computer Society Journals}
	%

\IEEEtitleabstractindextext{%
\begin{abstract}
  This paper studies the problem of cross-network node classification to overcome the insufficiency of labeled data in a single network. It aims to leverage the label information in a partially labeled source network to assist node classification in a completely unlabeled or partially labeled target network. Existing methods for single network learning cannot solve this problem due to the domain shift across networks. Some multi-network learning methods heavily rely on the existence of cross-network connections, thus are inapplicable for this problem. To tackle this problem, we propose a novel \textcolor{black}{graph} transfer learning framework AdaGCN by leveraging the techniques of adversarial domain adaptation and graph convolution. It consists of two components: a semi-supervised learning component and an adversarial domain adaptation component. The former aims to learn class discriminative node representations with given label information of the source and target networks, while the latter contributes to mitigating the distribution divergence between the source and target domains to facilitate knowledge transfer. Extensive empirical evaluations on real-world datasets show that AdaGCN can successfully transfer class information with a low label rate on the source network and a substantial divergence between the source and target domains. The source code for reproducing the experimental results is available at https://github.com/daiquanyu/AdaGCN.
\end{abstract}
		
\begin{IEEEkeywords}
  \textcolor{black}{Graph/Nework} Transfer Learning, Node Classification, Graph Convolution, Domain Adaptation, Adversarial Learning.
\end{IEEEkeywords}}
	
	\maketitle

	\IEEEdisplaynontitleabstractindextext

	%
	\IEEEpeerreviewmaketitle

\IEEEraisesectionheading{\section{Introduction}\label{sec:introduction}}
	
\IEEEPARstart{N}{ode} classification~\cite{social2011-BhagatCM11} is a central task in graph \textcolor{black}{(or network$\footnote{\textcolor{black}{In this paper, the terms graph and network are used interchangeably to denote graph-structured data.}}$)} analysis. It is an important building block of numerous real-world applications, such as product recommendation in e-commerce websites, advertisement distribution in social networks, and protein function identification for disease diagnosis. Many research efforts have been made to develop reliable and efficient methods for node classification in networked data.

In the era of big data, massive amount of raw data in information networks is produced everyday. However, labeled data is significantly expensive and slow to acquire due to the high cost and long time of human annotations, making it difficult to train a well-generalized classifier~\cite{corr-abs-1808-04572}. \textcolor{black}{Moreover, in some newly-formed networks, there may be no labels at all. It would be impossible to classify nodes with only the information of the network itself. To tackle these issues, a promising approach is to utilize class information from other similar or related networks to assist in classification, i.e., transfer learning on networked data~\cite{ICDM-FangYZ13,arxiv-shen-cdne}. For example, given a newly formed social network that is short of labels, to classify the users into different groups based on their interests, there is a need to utilize the abundant class information in some well-developed social networks. Moreover, in a newly collected protein-protein interaction network, to classify the proteins into different function categories, it would be beneficial to leverage the class information in a well-established protein database. Similarly, the class information in an early citation database can also be transferred to assign research topics for a newly constructed citation network.}

\begin{figure}[t]
	\centering
	\includegraphics[width=0.93\columnwidth]{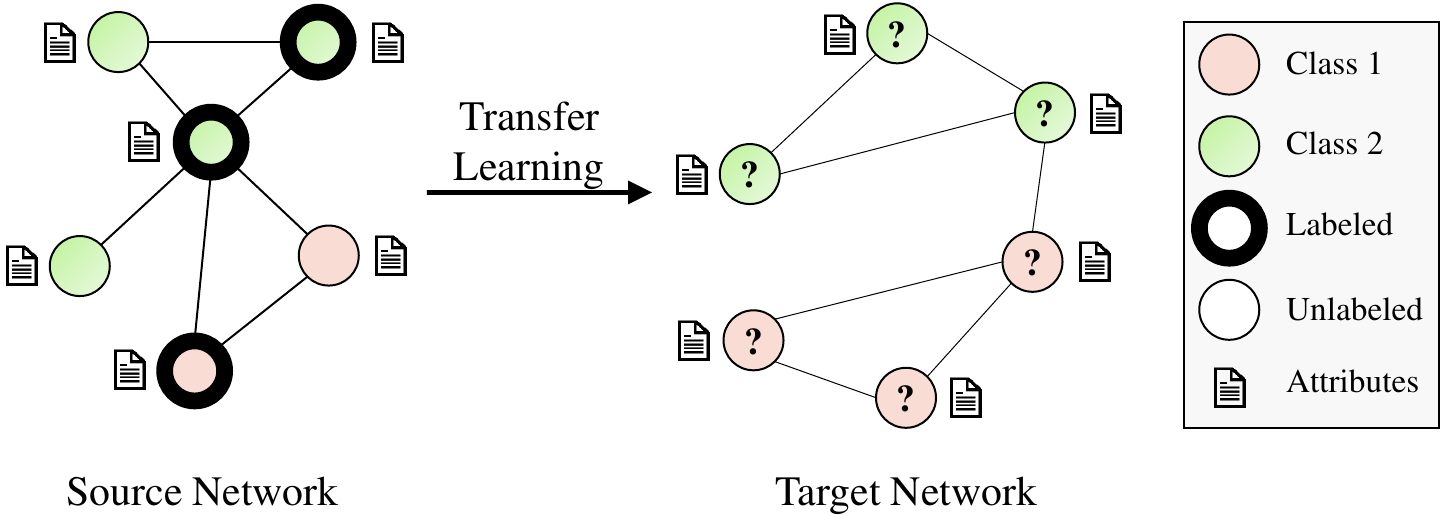}
	\caption{Cross-network node classification. We aim to transfer knowledge from a partially labeled source attributed network to assist the classification task in a completely unlabeled or partially labeled target attributed network. Here we use an unlabeled target network for illustration.}
	\label{problem_illustration}
	\vspace{-0.5em}
\end{figure}

In this paper, we consider a cross-network node classification problem that aims to leverage a partially labeled source attributed network to facilitate node classification in another completely unlabeled or partially labeled target attributed network (Figure~\ref{problem_illustration}). The challenges lie in several aspects. First, there may be a significant domain divergence between the source and target networks and they may not have many attributes in common. Second, there are no cross-network edges to propagate knowledge from the source network to the target network. Third, only a small portion of nodes in the source network are labeled.

Existing network embedding methods~\cite{KDD-14-Bryan,WWW-15-Jian,KDD-16-Grover} are insufficient to address these challenges. They first learn compact node representations to preserve network structural information, and then train a classifier with the learned representations for node classification. Most of these methods learn node representations in an unsupervised manner, and are often less effective than graph-based semi-supervised learning methods for node classification. Moreover, topology-only embedding methods cannot be easily generalized to cross-network problems due to lack of a similarity preserving component to push nodes of the same category from two networks close in the embedding space~\cite{MLG-17-HeimannKoutra}.
 
Graph-based semi-supervised learning methods ~\cite{ICML-16-YangCS,ICLR-16-KipfW} have been demonstrated highly effective for node classification in a single network with only a few labeled nodes. The recently proposed graph convolutional networks (GCN)~\cite{ICLR-16-KipfW} and follow-up works such as GraphSAGE~\cite{NIPS-17-HamiltonYL} and GAT~\cite{corr-abs-1710-10903}, naturally integrate network topology, node attributes and observed node labels into an end-to-end learning framework, and achieve superior performance on node classification. However, these methods are designed for learning tasks in a single network domain and will inherently have difficulties in generalizing to another network domain that may have a substantially different attribute set. 

There are some methods~\cite{WWW-NiCLCCX018,ICDM-FangYZ13} proposed to leverage the relationship between multiple networks to improve learning performance. Both EOE~\cite{WSDM-17-XuWCY} and DMNE~\cite{WWW-NiCLCCX018} learn embeddings for multiple networks simultaneously, but they heavily rely on the existence of cross-network connections, making them inapplicable for our problem. Currently there is little exploration of knowledge transfer across different networks for learning tasks such as node classification.

Domain adaptation utilizes the knowledge of relevant source domain(s) to assist the same learning task in the target domain~\cite{TKDE-PanY10,IJON-WangD18}. Although there are many existing domain adaptation methods for vector-based data such as images and texts (bag-of-words)~\cite{JMLR-GaninUAGLLML16,AAAI-ShenQZY18}, they are not applicable for graph-structured data, as entities in a graph are highly correlated with each other which violates the assumption of independent and identically distributed (IID) data samples in each individual domain. Little research has been conducted on domain adaptation for graph-structured data. CDNE~\cite{arxiv-shen-cdne} is the only attempt to our best knowledge, which learns transferable node embeddings for cross network learning tasks by minimizing the maximum mean discrepancy (MMD) loss. However, it cannot jointly model network structures and node attributes, which might limit its modeling capacity. Besides, it heavily relies on the preprocessing of the adjacency matrix with the positive pointwise mutual information (PPMI) matrix, which makes the sparse adjacency matrix denser and thus aggravates computational complexity due to the autoencoder-based model architecture.

To address the challenges for cross-network node classification, we propose a novel \textcolor{black}{graph} transfer learning framework AdaGCN that is based on \underline{\textbf{a}}dversarial \underline{\textbf{d}}omain \underline{\textbf{a}}daptation  with \underline{\textbf{g}}raph \underline{\textbf{c}}onvolutional \underline{\textbf{n}}etworks. The idea is to learn class discriminative node representations via graph convolutional networks and learn domain invariant node representations via adversarial learning. Hence, AdaGCN consists of a semi-supervised learning component and an adversarial domain adaptation component. 

On one hand, the semi-supervised component is dedicated to learning discriminative node representations for classification with the available labeled data from both the source and target networks. GCN enables training a well-behaved classifier with even only a small set of labeled nodes in the source network (as shown in Section~\ref{unsupervised}). However, the original GCN layer only conducts Laplacian smoothing on nearby nodes' features within one hop, and it requires stacking many layers to increase the smoothing level, which will greatly increase the number of trainable parameters and result in overfitting. To alleviate this issue, we propose to use an improved GCN layer designed with a smoothing strength hyperparameter\cite{CVPR-19-Qiamili}, which makes the model more efficient.

On the other hand, the adversarial domain adaptation component is aimed at mitigating the distribution shift between the source and target domains to encourage knowledge transfer by learning domain invariant representations via adversarial learning. Specifically, we model the domain adaptation process as a two-player game similar to GANs~\cite{NIPS-14-GoodfellowPMXWOCB}, where the representation learner GCN acts as the generator for learning domain invariant node representations while a domain critic as the discriminator is optimized to distinguish node representations from the source and target networks. By combining the two components, AdaGCN can learn both class discriminative and domain invariant node representations for transferring class information across networks.
 
Extensive experiments on real attributed networks show that AdaGCN can work in both unsupervised setting (i.e., completely unlabeled target network) and semi-supervised setting (i.e., scarcely labeled target network). Besides, it has low dependence on the common attributes shared by the source and target networks. The main contributions of this paper can be summarized as follows:
 
 \begin{itemize}[leftmargin=0.3cm]
  \item We pioneer in studying a challenging \textcolor{black}{graph} transfer learning problem under a realistic setting, where a partially labeled source network is utilized to assist node classification in a completely unlabeled or partially labeled target network.
  
  \item We develop a novel and principled framework for \textcolor{black}{graph} transfer learning by efficiently integrating techniques of adversarial domain adaptation and graph convolution.
  
  \item We conduct extensive experiments on real-world information networks to verify the effectiveness of our model, which demonstrates its superior performance compared with state-of-the-art baselines, impressive label efficiency, and good model robustness against distribution discrepancy.
 \end{itemize}

The organization of this paper is as follows. We review the literature in Section~\ref{related_workd}. We formulate the research problem in Section~\ref{problem_definition}. In Section~\ref{proposed_method}, a detailed description of the proposed methods is presented. Then, the experimental results and analysis are provided in Section~\ref{experiments}. Finally, a short summary with the contributions and
possible directions of future work are included in Section~\ref{conclusion}.
 
\section{Related Work} \label{related_workd}
\subsection{Single Network Learning}
Network embedding~\cite{survey-NE-18,survey-HamiltonYL17,survey-KG-21} is aimed at learning compact node representations based on network topology only or with side information in an unsupervised manner to facilitate a range of learning tasks, such as node classification and network visualization. For topology-only embedding methods, most of existing works focused on preserving network structures and properties in embedding vectors through various techniques such as negative sampling approach~\cite{WWW-15-Jian,KDD-14-Bryan,KDD-16-Grover}, matrix factorization technique~\cite{CIKM-15-SsCao,AAAI-17-XiaoW} and deep learning models~\cite{AAAI-16-SsCao,KDD-16-DxW,asunam-ShenC17,corr-xshen,TNNLS-TNE}. Most recently, regularization methods based on generative adversarial networks or adversarial training are exploited to handle noisy and incomplete networked data to improve generalization ability~\cite{AAAI-18-Quanyu,IJCAI-PanHLJYZ18,AdvT4NE_WWW2019,TCYB-PanHFLJZ20}. Aside from topology-only methods, many models are proposed to incorporate side information such as node attributes~\cite{IJCAI-ZhangYBZYZE018,IJCAI-PanHLJYZ18,WWW-XuWCY18,TNNLS-LYLZPSGZCKG21}. For example, ANRL~\cite{IJCAI-ZhangYBZYZE018} optimizes both network structure preserving loss and feature reconstruction loss based on stacked autoencoder. \textcolor{black}{LANE~\cite{WSDM-17-HuangLH} further incorporates label information in the attributed network embedding process.}

The unsupervised learning methods don't specially tailor the latent vectors for node classification, which makes them inferior to some customized models. Semi-supervised learning methods, including those using network topology and observed labels~\cite{IJCAI-16-TuZLS} and those combining network structures with available labels and node attributes~\cite{IJCAI-16-Pan,ICML-16-YangCS,ICLR-16-KipfW,NIPS-17-HamiltonYL,corr-abs-1710-10903,SDM-LiangJS018}, achieve state-of-the-art performance. Planetoid~\cite{ICML-16-YangCS} optimizes a supervised loss and a context-preserving loss. GCN~\cite{ICLR-16-KipfW} is a deep convolutional learning paradigm for graph-structured data which nicely integrates local node attributes and graph topology in convolutional layers. It further inspires lots of research work~\cite{NIPS-17-HamiltonYL,corr-abs-1710-10903,TKDD-WuPDZ21}. For example, GraphSAGE~\cite{NIPS-17-HamiltonYL} is a variant of GCN which designs different aggregation methods for feature extraction. GAT~\cite{corr-abs-1710-10903} improves GCN by leveraging attention mechanism to aggregate features from the neighbors of a node with discrimination.

While these methods can be modified to cross-network learning, the distribution drift between different network domains severely hampers knowledge transfer, especially for the topology-only methods~\cite{MLG-17-HeimannKoutra}.

\subsection{Multi-Network Learning}
A branch of work aims to leverage the relationship between multiple networks to facilitate learning, including those relying on inter-network edges~\cite{WSDM-17-XuWCY,WWW-NiCLCCX018}, those focusing on identifying common nodes across networks~\cite{IJCAI-LiuCLL16,CIKM-HeimannSSK18}, and those managing to transfer knowledge from the source network(s) to the target network(s)~\cite{WSDM-TangLK12,ICDM-FangYZ13,SIGIR-ShenCM17,Fuzzy-19-XiaoShen,arxiv-shen-cdne}. 

Both EOE~\cite{WSDM-17-XuWCY} and DMNE~\cite{WWW-NiCLCCX018} learn embeddings for multiple networks simultaneously. Specifically, EOE introduces a harmonious embedding matrix to model inter-network node similarities, while DMNE adapts autoencoder for multi-network embedding with a co-regularized loss to model cross-network relationships. However, these methods heavily rely on the existence of cross-network connections, which makes them inapplicable for our problem. Another line of related research is network alignment~\cite{IJCAI-LiuCLL16,CIKM-HeimannSSK18}, which aims to identify the node correspondence across networks with/without cross-network edges. It differs from our problem in the assumption of common nodes across networks and the research goal of finding common nodes across networks.

There is also some literature focusing on transferring knowledge from the source network(s) to the target network(s) for various tasks, such as social ties inference~\cite{WSDM-TangLK12}, positive/negative link prediction~\cite{WWW-YeCZC13}, and node classification~\cite{ICDM-FangYZ13,AAAI-LeeKLY17, fu2019nes}. In this paper, we aim to utilize knowledge in the source network to assist classification in the target network as~\cite{ICDM-FangYZ13,AAAI-LeeKLY17,arxiv-shen-cdne, fu2019nes}. In~\cite{ICDM-FangYZ13}, non-negative matrix factorization is jointly applied on the label propagation matrices of both the source and target networks so as to learn transferable structure features. However, it suffers from expensive computation in the matrix decomposition process, and it cannot jointly model the relationships among structural information, node attributes and node labels, which might cause negative transfer. CDNE~\cite{arxiv-shen-cdne} is closely related to our work. It first learns node embeddings for multiple networks with different stacked autoencoders and mitigates the distribution shift of node representations between networks by minimizing the MMD loss, and then trains a node classifier with the learned node representations. 
\textcolor{black}{NES-TL~\cite{fu2019nes} studies node popularity prediction on networks with multi-source transfer learning. 
It obtains node representations through feature engineering and employs an instance-based domain adaptation technique to reduce domain divergence. Our work differs from NES-TL in both the research problem and the techniques used. In this paper, we focus on solving the cross-network node classification problem by transferring information from a single source domain with graph convolution and adversarial domain adaptation techniques.}

\subsection{Domain Adaptation}
Domain adaptation is a subtopic of transfer learning, which aims to mitigate the harmful effect of domain drift when transferring knowledge from source to target~\cite{TKDE-PanY10,IJON-WangD18}. Approaches for domain adaptation can be classified into three groups, including the instance-based methods~\cite{AAAI-Tan0P017}, parameter-based methods~\cite{CORR-RozantsevSF16}, and feature-based methods~\cite{ICML-LongC0J15,ECCV-SunS16}. Among them, deep feature-based domain adaptation methods have attracted a lot of attention in recent years due to its effectiveness. They can be categoried into three branches, i.e., discrepancy-based methods~\cite{CORR-TzengHZSD14,ICML-LongC0J15}, reconstruction-based methods~\cite{IJCAI-ZhuangCLPH15,ICML-KimCKLK17}, and adversarial-based methods~\cite{JMLR-GaninUAGLLML16,CVPR-TzengHSD17,AAAI-ShenQZY18,aaai-PeiCLW18,nips-LongC0J18}.  

\textcolor{black}{In this paper, we are interested in adversarial-based methods for domain adaptation, which are motivated by the theory in~\cite{NIPS-Ben-DavidBCP06} and~\cite{ML-Ben-DavidBCKPV10}. It suggests that when the divergence between the representations of a source domain and a target domain is minimized, such representations would be good for knowledge transfer. 
The pioneering work DANN~\cite{JMLR-GaninUAGLLML16} learns domain invariant representations by formulating the problem as a minimax game similar to GANs~\cite{NIPS-14-GoodfellowPMXWOCB}. Specifically, 
a representation extractor, acting as the generator, is optimized to learn invariant representations for samples from the source and target domains. Meanwhile, a domain classifier, serving as the discriminator, is trained to tell apart the source and target representations by minimizing the domain classification loss.
Domain invariant representations can be obtained when the model converges and domain divergence is minimized.
To improve upon DANN, WDGRL~\cite{AAAI-ShenQZY18} employs the Wasserstein distance to quantify domain divergence, 
which leads to better gradient property and  generalization bound.
Besides, MADA~\cite{aaai-PeiCLW18} and CDAN~\cite{nips-LongC0J18} manage to leverage discriminative information from the label classifier to align the multi-modal distributions of representations from different domains. In this paper, we leverage adversarial-based techniques for domain adaptation on graph-structured data. The main difference is that the majority of previous methods are proposed for vector-based data such as image and text with the assumption of independent and identically distributed samples within each domain, whereas we investigate domain adaptation for graph-structured data with complicated correlations among data entities.}
 
\section{Problem Definition} \label{problem_definition}

\begin{table}
	\centering
	\label{notations}
	\caption{Notations}
	\renewcommand{\arraystretch}{1.2}
	\begin{adjustbox}{max width=0.46\textwidth}
		\begin{tabular}{ l  l }
			\hline
			Notation & Description \\
			\hline
			$G=(V, A, X)$ & A weighted attributed network \\
			$V$ & Node set of $G$ \\
			$A$ & Weighted adjacency matrix of $G$ \\
			$X$ & Feature matrix of $G$ \\
			$G^s=(V^s, A^s, X^s)$ & Source network \\
			$G^t=(V^t, A^t, X^t)$ & Target network \\
			$V^{sl}$ & Set of all labeled nodes in $G^s$ \\
			$Y^{sl}$ & Label matrix of $V^{sl}$ \\
			$\mathcal{X}^s$, $\mathcal{X}^t$ & Sets of node attributes of $G^s$ and $G^t$ \\
			$\mathcal{X}$ & Set of node attributes of both $G^s$ and $G^t$ \\
			$\mathcal{Y}$ & Label set of both $G^s$ and $G^t$\\
			\hline
			$n^s$, $n^t$ & \# of nodes in $G^s$ and $G^t$ \\
			$n^{sl}$ & \# of labeled nodes in $G^s$ \\
			$c^s$, $c^t$ & \# of node attributes in $G^s$ and $G^t$ \\
			$c$ & \# of node attributes in $\mathcal{X}$ \\
			$L$ & \# of categories in label space $\mathcal{Y}$ \\
			\hline
			$f_g$, $f_c$, $f_d$ & Representation learner, label classifier, and \\
			& domain critic \\
			$\boldsymbol{\theta}_g$, $\boldsymbol{\theta}_c$, $\boldsymbol{\theta}_d$ & Sets of model parameters in $f_g$, $f_c$, and $f_d$ \\
			$H^s_g$, $H^t_g$ & Source and target node representations \\
			$\lambda$ & Domain adaptation coefficient \\
			$\gamma$ & Gradient penalty coefficient \\
			$n_d$ & Domain critic training step per iteration\\
			$n_I$ & Smoothing parameter of the IGCN layer \\
			$\alpha_1$, $\alpha_2$ & Learning rates of domain critic, and \\
			& representation learner and label classifier \\
			\hline
		\end{tabular}
	\end{adjustbox}
	\label{tab-dataset}
	\vspace{-0.5em}
\end{table}

 In this paper, we study domain adaptation for networked data, i.e., leveraging the information of a source network to assist node classification in a completely unlabeled or partially labeled target network. The source network can be either partially labeled or fully labeled. In this section, we formally define the research problem and introduce notations used in the paper as summarized in Table 1.
 
 Denote by $G^s=(V^s, A^s, X^s)$ the source network, where $V^s$ is the node set ($n^s=|V^s|$), $A^s\in\mathbb{R}^{n^s\times n^s}$ is the weighted adjacency matrix with $A^s_{ij}$ quantifying the strength of connection between nodes $i$ and $j$, and $X^s\in\mathbb{R}^{n^s\times c^s}$ is the feature matrix with $c^s$ as the number of node attributes in $G^s$ and the $i$-th row of $X^s$ as the feature vector associated with node $i$. Denote by $V^{sl}$ the set of labeled nodes in $G^s$ and $Y^{sl}\in\mathbb{R}^{n^{sl}\times L}$ the label matrix of $V^{sl}$, where $Y^{sl}_{ik} = 1$ if node $i\in V^{sl}$ is associated with label $k$ and $Y^{sl}_{ik} = 0$ otherwise. 
 
 Similarly, the target network is represented as $G^t=(V^t, A^t, X^t)$, where $V^t$ is the node set $(n^t = |V^t|)$, $A^t\in\mathbb{R}^{n^t\times n^t}$ is the weighted adjacency matrix, and $X^t\in\mathbb{R}^{n^t\times c^t}$ is the feature matrix with $c^t$ as the number of node attributes in $G^t$. The target network $G^t$ can be either completely unlabeled or partially labeled. Here, we assume that it is completely unlabeled for simplicity, but our method can be straightforwardly extended to the partially labeled setting and we have conducted experiments for both scenarios in Section~\ref{RQ1} and~\ref{RQ2}.
 
 The source network and the target network may contain different attributes. Denote by $\mathcal{X}^s$ and $\mathcal{X}^t$ the set of node attributes in $G^s$ and $G^t$ respectively. We construct a new attribute set $\mathcal{X}=\mathcal{X}^s\cup \mathcal{X}^t$, where $c=|\mathcal{X}|$ represents the total number of attributes. We then reformulate the feature matrix of both $G^s$ and $G^t$ to make them include all the attributes in $\mathcal{X}$. With a slight abuse of notation, we still use $X^s\in \mathbb{R}^{n^s\times c}$ and $X^t\in \mathbb{R}^{n^t\times c}$ to represent the newly formed feature matrices of $G^s$ and $G^t$. In particular, $X_{ik}^r$ ($r\in\{s,t\}$) is the value of the $k$-th attribute associated with node $i$ in $G^r$ and $X^r_{ik}=0$ means that it is not associated with node $i$.
 
 Define a network domain as $D=\{G,f(G)\}$, which includes an attributed network $G$ and a function $f(G)$ for the node classification task. Then, the source network domain and the target network domain can be represented by $D^s=\{G^s,f(G^s)\}$ and $D^t=\{G^t,f(G^t)\}$, respectively. The problem considered in this paper is similar to the conventional domain adaptation problem as in~\cite{TKDE-PanY10,IJON-WangD18}. Specifically, there exists a domain divergence between the source and target networks, i.e., $D^s\neq D^t$, but the label space $\mathcal{Y}=\{1, \cdots, L\}$ is the same, and our goal is to learn a classifier $f$ to accurately classify the nodes in the target network with the assistance of the partially labeled source network.
 
\section{Proposed Method}\label{proposed_method}

\begin{figure}[t]
	\centering
	\includegraphics[width=0.95\columnwidth]{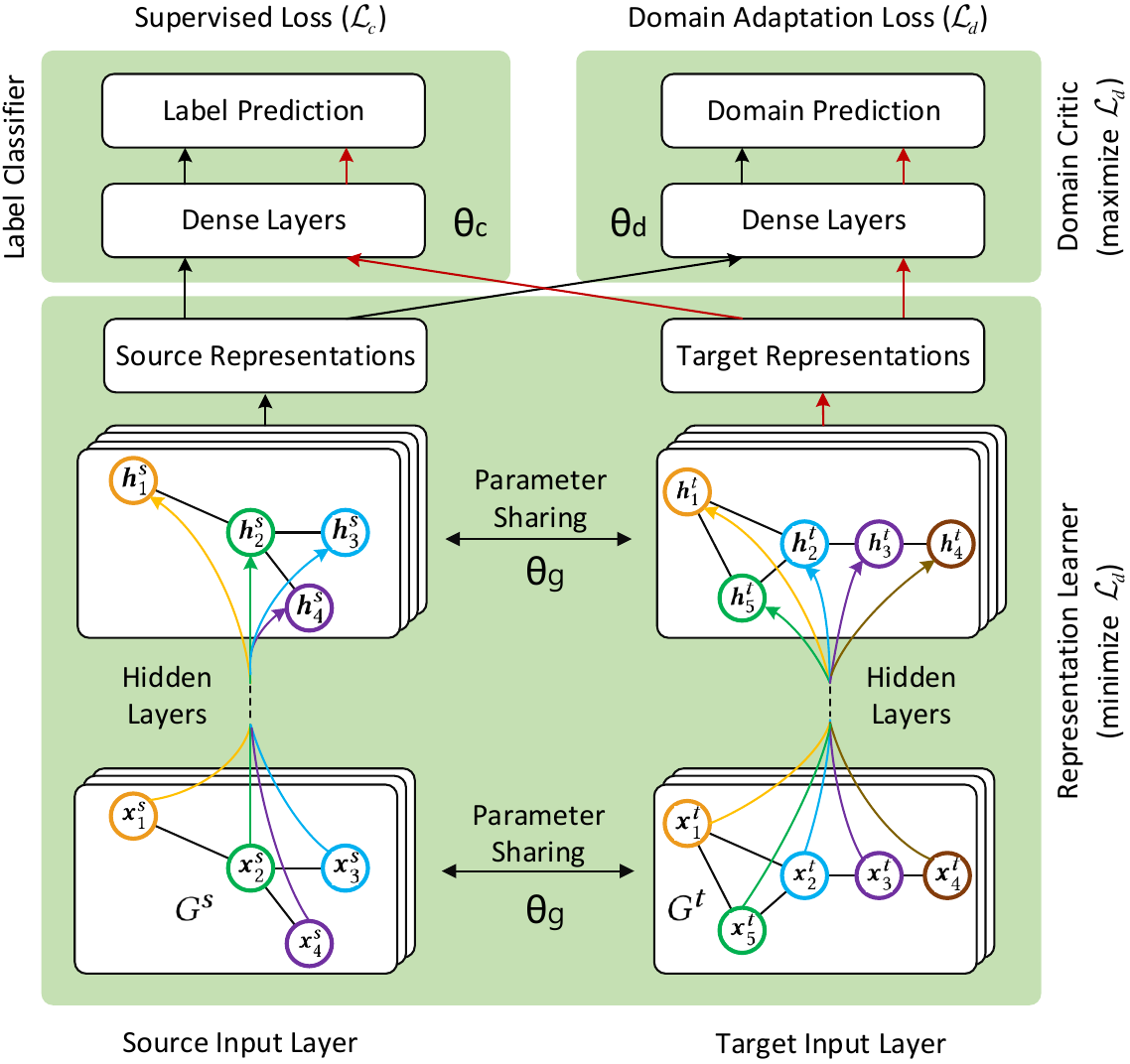}
	\caption{\textcolor{black}{Model architecture of AdaGCN. The representation learner computes representations for nodes in the source and target networks. The node representations are then fed to the label classifier and the domain critic for label predictions and domain predictions, respectively. The domain adaptation process is modeled as a minimax game between the representation learner and the domain critic.}}
	\label{gcn-Framework}
	\vspace{-0.5em}
\end{figure}

\subsection{An Overview of Model Architecture}
 
 To solve cross-network node classification problem, two major challenges need to be addressed. Firstly, how to fully exploit the available data information including graph structures, node attributes and observed node labels to learn useful node representations for the two networks? Secondly, how to overcome the serious domain divergence between two networks to facilitate knowledge transfer with the absence of cross network edges and only a few common node attributes across networks?
 
 To address the first challenge, we leverage graph convolution to integrate network topology and node attributes in a semi-supervised learning model, which is capable of learning discriminative node representations with available node labels. To tackle the second challenge, we manage to mitigate distribution discrepancy between two networks with the technique of adversarial domain adaptation. In particular, we propose a \textcolor{black}{graph} transfer learning framework AdaGCN by naturally combining the techniques of adversarial domain adaptation and graph convolution. The model architecture is shown in Figure~\ref{gcn-Framework}. \textcolor{black}{It consists of two components: a semi-supervised learning component, which includes a representation learner and a label classifier, and an adversarial domain adaptation component, which is composed of the representation learner and a domain critic. Therefore, the representation learner is shared by these two components.} With the cooperation of them, AdaGCN can learn both class discriminative and domain invariant node representations, thus enabling classifying nodes in the target network with only a few labeled nodes in the source network. Note that our model is also applicable for the semi-supervised scenario with a partially labeled target network.

\subsection{Network Representation Learning}

We propose to use graph convolution to jointly model network structures and node attributes for learning network representations, which has recently been demonstrated highly effective in various learning tasks such as node classification~\cite{ICLR-16-KipfW}, graph clustering~\cite{IJCAI-Xthlql19} and social recommendation~\cite{WWW-Fan0LHZTY19}.

Graph convolution is an operation that applies a linear graph convolutional filter \cite{shuman2013emerging,TSP-SandryhailaM13} $\hat{A}\in \mathbb{R}^{n\times n}$ on a graph signal $\boldsymbol h\in \mathbb{R}^{n}$ and outputs a new signal $\bar{\boldsymbol h}\in \mathbb{R}^{n}$ ($n$ is the number of nodes in the underlying graph):
\begin{equation}
  \bar{\boldsymbol h} = \hat{A}\boldsymbol h.
\end{equation}
The graph filter $\hat{A}$ is a matrix designed by manipulating the spectrum of the underlying graph. The graph signal $\boldsymbol h$ is a real-valued function defined on the nodes of the graph, i.e., each node is associated with a real number. For example, a column of the node feature matrix $X$ can be considered as a graph signal. 

Graph convolution provides a principled way to combine graph structures and node features for learning useful node representations. For the graph convolutional networks (GCN) proposed in \cite{ICLR-16-KipfW}, the graph filter is a renormalized adjacency matrix, which actually performs Laplacian smoothing that updates the features of each node with a weighted average of its own and neighbors' to obtain smooth embeddings~\cite{AAAI-LiHW18}. Further, it was shown in \cite{CVPR-19-Qiamili} that to produce smooth embeddings for nodes in the same cluster, the graph filter $\hat{A}$ needs to be low-pass. With a proper low-pass graph filter, graph convolution will generate useful representations that help to ease knowledge transfer across networks and node classification in the target network. 
   
In this paper, we propose two methods for learning network representation with graph convolution. The first one is based on the layer-wise propagation rule of GCN. Specifically, the hidden representations of the $k$-th convolutional layer in the feature extractor are learned by:
\begin{equation}
H_g^{(k)} = \sigma(\hat{A}H_g^{(k-1)}W_g^{(k)}),
\label{gcn_hidden}
\end{equation}
where $\hat{A}$ is a renormalized adjacency matrix with a self-loop at each
node, $H_g^{(k-1)}$ is the output of the previous layer ($H_g^{0}=X$), $W_g^{(k)}$ is a projection matrix with trainable parameters, and $\sigma(\cdot)$ is the activation function. As illustrated in Figure~\ref{gcn-Framework}, we use two GCNs for learning node representations for the source and target networks respectively, but they share a common set of trainable parameters ($W_g^{(k)}$) so as to help transfer knowledge across networks. For simplicity of notation, we denote a GCN as $f_g(A, X;\boldsymbol{\theta}_g)$, which takes the graph adjacency matrix $A$ and the feature matrix $X$ as input, and $\boldsymbol{\theta}_g$ represents the trainable parameters. Then, we can obtain the output node representations of the source and target networks as:
\begin{equation}
\begin{aligned}
H_g^r = f_g(A^r, X^r;\boldsymbol{\theta}_g),\;r\in\{s, t\},
\end{aligned}
\end{equation}
where $[H_g^r]_i$ as the $i$-th row of $H_g^r$ is the representation of node $i$.

However, with the GCN layer defined as in Eq.~(\ref{gcn_hidden}), one has to stack multiple layers to increase the strength of feature smoothing, which will also increase model complexity because of the accompanied trainable parameters in each layer, and thus can easily cause overfitting, especially for learning tasks with low source training rates. To address this issue, we propose to use an improved GCN (IGCN) layer proposed in \cite{CVPR-19-Qiamili} to improve the strength of the graph convolutional filter for learning better representations. Then, for our second method, the hidden representations of the $k$-th convolutional layer are obtained with: 
\begin{equation}
H_g^{(k)} = \sigma({\hat{A}}^{n_I}H_g^{(k-1)}W_g^{(k)}),
\label{igcn_hidden}
\end{equation}
where $n_I$ is the exponent of $\hat{A}$, i.e., the smoothing parameter. By setting an appropriate $n_I$, we can easily control the smoothing strength of graph convolution to facilitate knowledge transfer and classification while avoiding overfitting. As suggested in \cite{CVPR-19-Qiamili}, normally, larger $n_I$ should be used with lower source training rates.
 
\subsection{Semi-Supervised Learning}
 In AdaGCN, the node representations of the source and target networks learned by GCNs will be fed to a classifier for label prediction, and together they form the semi-supervised learning component. The classifier could be a single layer logistic regression classifier or a multi-layer perceptron. We denote the classifier as $f_c(H;\boldsymbol{\theta}_c)$, where $H$ represents the node representations as input and $\boldsymbol{\theta}_c$ represents the trainable parameters. We then denote the prediction scores of nodes in the source and target networks as:
 \begin{equation}
 \hat{Y}^r = f_c(H^r_g;\boldsymbol{\theta}_c),\;r\in\{s, t\},
 \label{label_classifier}
 \end{equation}
 where $H^r_g$ are the node representations generated by GCNs and $\hat{Y}^r_{ik}$ $(r\in\{s, t\})$ represents the prediction score for node $i$ in class $k$. One can conduct multi-class or multi-label classification by changing the activation function of the output layer in the classifier $f_c(\cdot)$. For multi-class classification, the activation function can be the softmax function. For multi-label classification, the activation function is the sigmoid function. \textcolor{black}{We use the cross-entropy loss over all the labeled nodes in the source network as the objective for multi-label classification:
 \begin{equation}
 \mathcal{L}_c = -\frac{1}{n^{sl}}\sum\limits_{i=1}^{n^{sl}}\sum\limits_{k=1}^{L} \left[Y^{sl}_{ik}\log(\hat{Y}^{sl}_{ik})+\left(1-Y^{sl}_{ik}\right)\log\left(1-\hat{Y}^{sl}_{ik}\right)\right],
 \end{equation}
 where $\hat{Y}^{sl}$ is the prediction score matrix of the labeled nodes $V^{sl}$ in the source network. Note that our method can be easily extended to the semi-supervised setting by incorporting available target labels into the above loss.}
 
\subsection{Adversarial Domain Adaptation}
 \textcolor{black}{The domain adaptation theory in~\cite{NIPS-Ben-DavidBCP06} and~\cite{ML-Ben-DavidBCKPV10} suggests that, when the divergence between the representations of the source domain and those of the target domain is minimized, it is possible to transfer knowledge from the source network to the target network.
 In AdaGCN, we leverage the adversarial domain adaptation method~\cite{JMLR-GaninUAGLLML16,AAAI-ShenQZY18} to achieve this. Specifically, we model the domain adaptation process as a two-player game similar to GANs~\cite{NIPS-14-GoodfellowPMXWOCB}, where a representation learning network $f_g(A,X;\boldsymbol{\theta}_g)$ is acting as the generator for learning network invariant node representations, while a domain critic acting as the discriminator is optimized to distinguish node representations from the source and target networks. After adversarial training, the representation learning network is expected to generate similar representations for the source and target networks. Therefore network invariant representations can be obtained, and class information can be transferred from the source network to the target network. This is empirically validated by experimental results in Section~\ref{experiments}, where the proposed method AdaGCN demonstrates superiority over GCN, its counterpart without domain adaptation.}
 
 In the original GANs~\cite{NIPS-14-GoodfellowPMXWOCB}, the domain critic is a binary classifier, and the generator and the discriminator fight against each other over a log likelihood objective. However, directly formulating the problem as a binary classification problem and leveraging cross-entropy loss for model optimization may suffer from training instability such as mode collapse~\cite{ICML-ArjovskyCB17,NIPS-GulrajaniAADC17}. To improve learning stability, we instead minimize the Wasserstein-1 distance between the source and target distributions of node representation as suggested in~\cite{ICML-ArjovskyCB17,NIPS-GulrajaniAADC17,AAAI-ShenQZY18}. 
 
 We set the domain critic as a fully-connected neural network that takes a node representation as input and returns a real number. Denote by $f_d(\boldsymbol{h};\boldsymbol{\theta}_d)$ the domain critic, where $\boldsymbol{h}=[f_g(A,X;\boldsymbol{\theta}_g)]_v$ is the representation of node $v$ generated by a GCN with $X$ as the input node feature matrix, and $\boldsymbol{\theta}_d$ represents the trainable parameters. The first Wasserstein distance between the source and target distributions of node representation $\mathbb{P}_{\boldsymbol{h}^s}$ and $\mathbb{P}_{\boldsymbol{h}^t}$ can be computed using the Kantorovich-Rubinstein duality~\cite{opt_transport}:
 \begin{equation}
 W_1(\mathbb{P}_{\boldsymbol{h}^s}, \mathbb{P}_{\boldsymbol{h}^t}) 
 = \sup\limits_{\parallel f_d\parallel_{L_c}\leq 1} \mathbb{E}_{\mathbb{P}_{\boldsymbol{h}^s}}[f_d(\boldsymbol{h};\boldsymbol{\theta}_d)]-\mathbb{E}_{\mathbb{P}_{\boldsymbol{h}^t}}[f_d(\boldsymbol{h};\boldsymbol{\theta}_d)],
 \label{wgan_loss}
 \end{equation}
 where $\parallel f_d\parallel_{L_c}\leq 1$ is the Lipschitz continuity constraint. It can be interpreted as the minimum cost of transporting mass for transforming one distribution into another with the cost defined as the mass times the transport distance~\cite{ICML-ArjovskyCB17}. We can further approximate the empirical Wasserstein distance under the 1-Lipschitz assumption by maximizing the following domain critic loss with respect to $\boldsymbol{\theta}_d$:
 \begin{equation}
 \begin{array}{ll}
 \mathcal{L}_{d} = \frac{1}{n^s}\sum\limits_{i=1}^{n^s}f_d([f_g(A^s,X^s; \boldsymbol{\theta}_g)]_i;\boldsymbol{\theta}_d) \\ \quad\;\; -\frac{1}{n^t}\sum\limits_{i=1}^{n^t}f_d([f_g(A^t,X^t, \boldsymbol{\theta}_g)]_i;\boldsymbol{\theta}_d).
 \end{array}
 \end{equation}
 To enforce the Lipschitz constraint, we add a gradient penalty $\mathcal{L}_{grad}$ for the parameters $\boldsymbol{\theta}_d$ of the domain critic as suggested in~\cite{NIPS-GulrajaniAADC17}:
 \begin{equation}
 \mathcal{L}_{grad}(\hat{\boldsymbol{h}}) = (\parallel \nabla_{\hat{\boldsymbol{h}}}f_d(\hat{\boldsymbol{h}};\boldsymbol{\theta}_d)\parallel_2-1)^2,
 \end{equation}
 where the representation $\hat{\boldsymbol{h}}$ can be the source representations, the target representations, and the random points along the straight line between the source and target representation pairs. It can help avoid the capacity underuse and gradient vanishing/exploding problems of weight clipping methods~\cite{ICML-ArjovskyCB17} for 1-Lipschitz enforcement.  
  
 Hence, we solve the following minimax problem for learning network invariant node representations: 
 \begin{equation}
 \min\limits_{\boldsymbol{\theta}_g}\max\limits_{\boldsymbol{\theta}_d} \{\mathcal{L}_{d}-\gamma\mathcal{L}_{grad}\},
 \end{equation}
 where $\gamma$ is the gradient penalty coefficient, which should be set to 0 when optimizing the generator. The optimization problem suggests that the domain critic $f_d(\cdot)$ should be first trained to be optimal and then parameters in the generator $f_g(\cdot)$ are updated to minimize the Wasserstein distance between the source and target node representations.
 
 Note that our proposed AdaGCN is very flexible, and some other adversarial-based domain adaptation methods~\cite{aaai-PeiCLW18,nips-LongC0J18} can also be integrated into our framework.

\begin{algorithm}[t]\small
	\caption{Training Algorithm of AdaGCN}
	\label{gcn_da_algorithm}
	\SetKwInOut{Input}{Input}
	\SetKwInOut{Output}{Output}
	
	\Input{source data $\{G^s=(V^s, A^s, X^s), Y^{sl}\}$, target data $\{G^t=(V^t, A^t, X^t)\}$, domain critic training step $n_d$, coefficients $\gamma$, $\lambda$, learning rates $\alpha_1$, $\alpha_2$}
	Initialize parameters $\boldsymbol{\theta}_g$ for representation learner $f_g$, $\boldsymbol{\theta}_c$ for label classifier $f_c$, and $\boldsymbol{\theta}_d$ for domain critic $f_d$\;
	\While{not converge}{
		$//$ Optimize domain critic \\
		\For{$t=1,\dots,n_d$}{
			$H^s_g\leftarrow f_g(A^s,X^s;\boldsymbol{\theta_g})$, $H^t_g\leftarrow f_g(A^t,X^t;\boldsymbol{\theta_g})$\; $N\leftarrow \min\{N^s, N^t\}$\;
			Construct $H=\{\boldsymbol{h}_i\}_{i=1}^N$ with $\boldsymbol{h}_i\leftarrow \varepsilon \boldsymbol{h}^s+(1-\varepsilon)\boldsymbol{h}^t$, where $\epsilon$ is a random number sampled from $U[0,1]$, $\boldsymbol{h}^s$ and $\boldsymbol{h}^t$ are sampled from $H^s_g$ and $H^t_g$, respectively\;
			$\hat{H}\leftarrow \{H^s_g, H^t_g, H\}$\; 
			$\boldsymbol{\theta}_d\leftarrow \boldsymbol{\theta}_d+\alpha_1\cdot \nabla_{\boldsymbol{\theta}_d}\{\mathcal{L}_{d}-\gamma\mathcal{L}_{grad}(\hat{H})\}$\;
		}
		$//$ Optimize representation learner and label classifier\\
		$\boldsymbol{\theta}\leftarrow \{\boldsymbol{\theta}_g, \boldsymbol{\theta}_c\}$\;
		$\boldsymbol{\theta}\leftarrow \boldsymbol{\theta} - \alpha_2\cdot \nabla_{\boldsymbol{\theta}}\{\mathcal{L}_c+\lambda\mathcal{L}_{d}\}$\;
	}
\end{algorithm}

 \subsection{Overall Loss and Model Training}
 The overall loss of the proposed model AdaGCN is as follows:
 \begin{equation}
     \min\limits_{\boldsymbol{\theta}_g, \boldsymbol{\theta}_c}\{\mathcal{L}_c+\lambda\max\limits_{\boldsymbol{\theta}_d}[\mathcal{L}_{d}-\gamma\mathcal{L}_{grad}]\},
 \end{equation}
 where $\lambda$ is the coefficient for balancing semi-supervised learning and domain adaptation. We summarize the training procedure for AdaGCN in Algorithm~\ref{gcn_da_algorithm}. Note that here we do a full-batch training with gradient descent, but some existing methods can be applied to train the model in a mini-batch manner~\cite{ICML-ChenZS18,ICLR-jchentfm18}. First, as presented in line 4-10, we optimize the parameters $\boldsymbol{\theta}_d$ of the domain critic $f_d(\cdot)$ via gradient descent with other model parameters fixed. Then, as shown in lines 12 and 13, we fix $\boldsymbol{\theta}_d$, and update the parameters $\boldsymbol{\theta}_g$ of the generator $f_g(\cdot)$ and $\boldsymbol{\theta}_c$ of the classifier $f_c(\cdot)$ by minimizing the classification loss $\mathcal{L}_c$ and the domain adaptation loss $\mathcal{L}_d$ simultaneously. When the model converges, we can obtain class discriminative and domain invariant node representations. To classify nodes in the target network, one can simply feed the learned node representations to the trained classifier $f_c(\cdot)$.
 
\textcolor{black}{\subsection{Time Complexity Analysis}}
\textcolor{black}{The computational complexity of the model mainly consists of three parts: the GCN layers (Eq.~(\ref{gcn_hidden})), the label classifier (Eq.~(\ref{label_classifier})), and the domain critic (Eq.~(\ref{wgan_loss})).} It takes $\mathcal{O}((|E^s|+|E^t|)w_1w_2)$ (suppose that $E^s$ and $E^t$ are the edge sets of the source and target networks respectively, and $W_g^{(k)}\in \mathbb{R}^{w_1\times w_2}$) to compute the hidden representations with a single GCN layer for both the source and target networks through Eq.~(\ref{gcn_hidden}), which is linear to the number of edges. Note that the IGCN layer can ensure linearity with only an additional constant scale factor $n_I$ added to the complexity through left multiplying $H_g^{(k)}$ by $\hat{A}$ repeatedly for $n_I$ times in Eq.~(\ref{igcn_hidden}). Obviously, the time complexity of the label classifier or the domain critic is linear to the number of nodes. Thus, the overall complexity of the proposed methods is linear to the size of the networks.

\section{Experiments} \label{experiments}
 In this section, we aim to answer the following research questions (RQ) via experiments:
 \begin{description}[style=unboxed,leftmargin=0.55cm]
    \item[RQ1] How do the proposed methods perform compared with state-of-the-art methods? 
    \item[RQ2] How do the training rates of the source and target networks, i.e., the ratio of labeled nodes in $G^s$ and $G^t$, affect the performance of transfer learning? 
    \item[RQ3] How does the distribution discrepancy between the source and target networks affect the results of transfer learning? 
    \item[RQ4] How does the strength of graph convolution affect the domain adaptation performance? 
    \item[RQ5] How do the hyper-parameters affect the performance of the proposed methods? 
 \end{description}
 We also visualize the learned node embeddings from representation learner to provide an intuitive understanding of our proposed methods.

\subsection{Experiment Setup}
\subsubsection{Datasets}

\begin{table}
	\renewcommand{\arraystretch}{1.4}
	\centering
	\caption{Statistics of the real-world network datasets}
	\vspace{-0.5em}
	\begin{adjustbox}{max width=0.46\textwidth}
		\begin{tabular}{  c  c  c  c  c  c }
			\hline
			Dataset & \#Nodes & \#Edges & \#Attributes & \#Union Attributes & \#Labels \\
			\hline
			DBLPv7 & 5,484 & 8,130 & 4,412 & \multirow{3}{*}{6,775} & \multirow{3}{*}{5}\\
			Citationv1 & 8,935 & 15,113 & 5,379 &&\\
			ACMv9 & 9,360 & 15,602 & 5,571 &&\\
			\hline
		\end{tabular}
	\end{adjustbox}
	\label{tab-dataset}
	\vspace{-0.5em}
\end{table}

 We conduct experiments on three real-world attributed networks constructed by~\cite{arxiv-shen-cdne} based on datasets provided by ArnetMiner~\cite{KDD-08-TangZYLZS}. Some statistics of the experimental datasets are displayed in Table~\ref{tab-dataset}. DBLPv7, Citationv1 and ACMv9 are three paper citation networks from different original sources, i.e., DBLP, Microsoft Academic Graph and ACM respectively, and contain papers published in different periods, i.e., between years 2004 and 2008, before year 2008, and after year 2010, respectively. Here we consider them as undirected networks with each edge representing a citation relation between two papers. Each paper belongs to some of the following five categories according to its research topics, including ``Databases'', ``Artificial Intelligence'', ``Computer Vision'', ``Information Security'', and ``Networking''. Besides, the keywords extracted from the title of each paper were utilized as its attributes in the form of bag-of-words vector. We evaluate our proposed methods by conducting multi-label classification on these three network domains through six transfer learning tasks including C$\rightarrow$D, A$\rightarrow$D, D$\rightarrow$C, A$\rightarrow$C, D$\rightarrow$A, and C$\rightarrow$A, where D, C, A denote DBLPv7, Citationv1 and ACMv9, respectively.
 
\subsubsection{Baselines}
 We select baselines from several related research lines including single network embedding methods, graph-based semi-supervised learning methods, deep domain adaptation methods, and transfer learning methods for networked data. The descriptions of them are listed as follows:

 \begin{itemize}[leftmargin=0.3cm]
    \item \textbf{DeepWalk}~\cite{KDD-14-Bryan}, \textbf{node2vec}~\cite{KDD-16-Grover}: They are single network embedding methods using network structure only. Both DeepWalk and node2vec first transform network topology into node sequences, and then use skip-gram model to learn node representations. 
    \item  \textcolor{black}{\textbf{ANRL}~\cite{IJCAI-ZhangYBZYZE018}, \textbf{LANE}~\cite{WSDM-17-HuangLH}: They are attributed network embedding methods. ANRL is a deep model adapted from autoencoder, and we use its best variant ANRL-WAN. LANE jointly projects an attributed network and its node labels into a unified embedding space by eigenvector decomposition.}
    \item \textbf{GCN}~\cite{ICLR-16-KipfW}, \textbf{GraphSAGE}~\cite{NIPS-17-HamiltonYL}: They can be used for semi-supervised learning and representation learning.GCN is a deep convolutional network for graph-structured data, which integrates network topology, node attributes and observed labels into an end-to-end learning framework. GraphSAGE is a variant of GCN with different aggregation methods.
    \item \textbf{DNNs}, \textbf{WDGRL}~\cite{AAAI-ShenQZY18}: These two deep models only utilize node attributes. DNNs is a multi-layer perceptron. WDGRL is a state-of-the-art adversarial domain adaptation method with the assumption of IID vector-based inputs in each domain.
    \item \textbf{NetTr}~\cite{ICDM-FangYZ13}, \textbf{CDNE}~\cite{arxiv-shen-cdne}: These are transfer learning methods for networked data. NetTr learns transferable representations based on network topology only. CDNE is adapted from autoencoder by adding MMD loss across networks for domain adaptation.
 \end{itemize}
 We denote our methods with regular GCN layers (Eq.~(\ref{gcn_hidden})) as AdaGCN and improved GCN layers (Eq.~(\ref{igcn_hidden})) as AdaIGCN.

\subsubsection{Implementation Details}\label{parameter_settings}
 We implement our proposed methods using Tensorflow with Adam optimizer. For all transfer learning tasks, we use the same set of parameter configurations unless otherwise specified. We first describe the settings of AdaGCN. The GCNs of both the source and target networks contain three hidden layers with structure as 1000-100-16. The dropout rate for each GCN layer is set to 0.3. The classifier $f_c(\cdot)$ is a logistic regression model with sigmoid output layer for multi-label classification. The domain critic $f_d(\cdot)$ contains only one hidden layer with 16 units. A $l_2$-norm regularization term is imposed on model parameters except those of $f_d(\cdot)$ with the regularization coefficient as $5\times 10^{-5}$. The domain adaptation coefficient $\lambda$, gradient penalty coefficient $\gamma$, and domain critic training step $n_d$ are set to 1, 10 and 10, respectively. The learning rates for both components of our method are set to $1.5\times 10^{-3}$. We train the model for 1000 epochs, and perform a learning rate decaying by multiplying a decaying factor 0.8 per 100 epochs after the first 500 epochs to stabilize training. For AdaIGCN, it has similar configurations as AdaGCN with the only difference in the representation learner, which consists of only one IGCN layer and two additional fully connected layers. $n_I$ is set to 10 for all tasks. GCN and AdaGCN have the same settings for common hyper-parameters and model structure.
 
 For single network embedding methods, including DeepWalk, node2vec, and ANRL, node representations are first learned and then a one-vs-rest logistic regression classifier is trained with labeled nodes of both networks. For fair comparison, the dimension of node representations for these methods are all set to 128. \textcolor{black}{LANE and GraphSAGE are implemented based on the source codes provided by the authors.} For GraphSAGE, we adapt it to the transductive setting for better utilization of linkage information of the two networks, and use its best variant GraphSAGE-LSTM for comparison. Since these methods are designed for single network, we simply combine two networks into one and then conduct experiments as single network learning. In such combined networks, there are no any edge connections between source and target networks. 
 
 DNNs have similar parameter settings with GCN, and WDGRL have similar parameter settings with AdaGCN. We have also tried to improve the input features of DNNs and WDGRL by augmenting the feature matrix of graph with the learned embedding vectors from DeepWalk, but found it deteriorates performance, which is explainable since the learned embeddings of the source and target networks from DeepWalk are not comparable. Experiments for NetTr and CDNE are conducted following the original papers.

\begin{table*}[t]
	\renewcommand{\arraystretch}{1.4}
	\caption{Multi-label classification with source training rate as 10\%}
	\centering
	\scalebox{0.85}{
	\centerline{
	\begin{threeparttable}
			\begin{tabular}
				{ccccccccccccc|cc}
				\hline
				\multicolumn{1}{l}{} & Source & Target & DeepWalk & node2vec & ANRL & \textcolor{black}{LANE} & GraphSAGE & DNNs & WDGRL & NetTr & CDNE & GCN & AdaGCN & AdaIGCN \\
				\hline
				\multirow{6}{*}{\tabincell{c}{Micro-F1\\(\%)}} & C & \multirow{2}{*}{D} 
				& 24.67 & 21.73 & 49.37 & \textcolor{black}{53.30} & 67.42 & 28.71 & 29.05 & 48.32 & 70.80 & 67.00 & \textbf{70.89} & \textbf{75.14} \\ & A &
				& 25.96 & 27.70 & 48.41 & \textcolor{black}{50.37} & 65.95 & 33.88 & 32.86 & 48.24 & 62.94 & 65.92 & \textbf{69.32} & \textbf{74.85} \\ \cline{2-15} & D & \multirow{2}{*}{C}
				& 31.05 & 21.78 & 40.47 & \textcolor{black}{47.86} & 59.23 & 16.36 & 21.63 & 44.47 & 71.34 & 63.47 & \textbf{77.77} & \textbf{79.34} \\ & A &
				& 25.34 & 23.48 & 46.55 & \textcolor{black}{49.57} & 66.49 & 27.33 & 28.82 & 47.60 & 72.10 & 69.02 & \textbf{78.83} & \textbf{78.97} \\ \cline{2-15} & D & \multirow{2}{*}{A}
				& 27.85 & 26.24 & 39.99 & \textcolor{black}{47.02} & 53.08 & 9.64  & 16.90 & 42.42 & 66.79 & 53.32 & \textbf{67.92} & \textbf{71.43} \\ & C &  
				& 28.98 & 19.39 & 44.22 & \textcolor{black}{51.01} & 61.05 & 24.63 & 28.02 & 44.73 & \textbf{71.02} & 64.33 & 68.30 & \textbf{74.48} \\ \cline{2-15}  & \multicolumn{2}{c}{\textcolor{black}{Average}}  
				& \textcolor{black}{27.31} & \textcolor{black}{23.39} & \textcolor{black}{44.84} & \textcolor{black}{49.86} & \textcolor{black}{62.20} & \textcolor{black}{23.43} & \textcolor{black}{26.21} & \textcolor{black}{45.96} & \textcolor{black}{69.17} & \textcolor{black}{63.84} & \textcolor{black}{\textbf{72.17}} & \textcolor{black}{\textbf{75.70}} \\ 
				\hline
				\multirow{6}{*}{\tabincell{c}{Macro-F1\\(\%)}} & C & \multirow{2}{*}{D} 
				& 22.02 & 15.32 & 43.19 & \textcolor{black}{45.41} & 62.03 & 28.14 & 29.27 & 42.63 & 68.62 & 64.28 & \textbf{69.75} & \textbf{72.53} \\ & A &  
				& 21.90 & 23.53 & 40.11 & \textcolor{black}{41.87} & 61.66 & 31.95 & 31.79 & 41.78 & 60.87 & 62.24 & \textbf{68.46} & \textbf{72.29} \\ \cline{2-15} & D & \multirow{2}{*}{C} 
				& 25.14 & 16.50 & 34.13 & \textcolor{black}{40.32} & 52.13 & 16.61 & 21.33 & 39.66 & 70.47 & 59.76 & \textbf{76.42} & \textbf{77.95}\\ & A &  
				& 20.94 & 19.70 & 40.58 & \textcolor{black}{42.77} & 61.54 & 26.83 & 28.36 & 42.88 & 70.29 & 65.10 & \textbf{77.45} & \textbf{77.53} \\ \cline{2-15} & D & \multirow{2}{*}{A} 
				& 25.86 & 20.90 & 33.32 & \textcolor{black}{38.37} & 45.85 & 10.11 & 17.17 & 36.17 & 65.72 & 50.12 & \textbf{69.31} & \textbf{72.26} \\ & C &  
				& 23.85 & 14.38 & 38.94 & \textcolor{black}{43.95} & 52.98 & 24.43 & 27.39 & 40.78 & \textbf{70.16} & 61.35 & 67.83 & \textbf{75.04} \\ \cline{2-15} & \multicolumn{2}{c}{\textcolor{black}{Average}}  
				& \textcolor{black}{23.29} & \textcolor{black}{18.39} & \textcolor{black}{38.38} & \textcolor{black}{42.12} & \textcolor{black}{56.03} & \textcolor{black}{23.01} & \textcolor{black}{25.89} & \textcolor{black}{40.65} & \textcolor{black}{67.69} & \textcolor{black}{60.48} & \textcolor{black}{\textbf{71.54}} & \textcolor{black}{\textbf{74.60}} \\
				\hline
		\end{tabular}
		\begin{tablenotes}\footnotesize
			\item[*] D: DBLPv7, C: Citationv1, A: ACMv9. The top 2 classification f1-scores are highlighted in bold for each task.
		\end{tablenotes}
	\end{threeparttable}}}
	\label{source_training_rate_10}
	\vspace{-0.5em}
\end{table*}

\begin{table*}[t]
	\renewcommand{\arraystretch}{1.4}
	\caption{Multi-label classification with source training rate as 10\% and target training rate as 5\%}
	\centering
	\scalebox{0.85}{
	\centerline{
	\begin{threeparttable}
			\begin{tabular}{ccccccccccccc|cc}
				\hline
				\multicolumn{1}{l}{} & Source & Target & DeepWalk & node2vec & ANRL & \textcolor{black}{LANE} & GraphSAGE & DNNs & WDGRL & NetTr & CDNE & GCN & AdaGCN & AdaIGCN \\ 
				\hline
				\multirow{6}{*}{\begin{tabular}[c]{@{}c@{}}Micro-F1\\ (\%)\end{tabular}} & C & \multirow{2}{*}{D} & 53.10 & 59.17 & 55.97 & \textcolor{black}{56.73} & 70.42 & 32.41 & 33.59 & 50.74 & \textbf{73.69} & 71.25 & 71.90 & \textbf{75.25} \\ 
				& A &  & 48.02 & 57.67 & 52.55 & \textcolor{black}{55.89} & 69.13 & 40.18 & 30.73 & 48.31 & 69.61 & 70.29 & \textbf{75.18} & \textbf{75.95} \\ \cline{2-15} 
				& D & \multirow{2}{*}{C} & 66.57 & 69.13 & 53.91 & \textcolor{black}{58.12} & 68.99 & 28.03 & 23.33 & 50.28 & 78.93 & 73.09 & \textbf{79.66} & \textbf{80.33} \\ 
				& A &  & 61.56 & 66.91 & 54.42 & \textcolor{black}{58.76} & 71.92 & 34.60 & 32.73 & 49.98 & 77.86 & 75.28 & \textbf{81.45} & \textbf{82.00} \\ \cline{2-15} 
				& D & \multirow{2}{*}{A} & 58.85 & 64.23 & 49.37 & \textcolor{black}{54.52} & 64.64 & 27.57 & 21.76 & 45.24 & \textbf{77.38} & 71.51 & 75.15 & \textbf{78.18} \\ 
				& C &  & 57.58 & 62.60 & 51.39 & \textcolor{black}{55.56} & 69.20 & 35.17 & 33.43 & 46.26 & \textbf{77.10} & 72.84 & 74.51 & \textbf{77.14} \\ \cline{2-15} & \multicolumn{2}{c}{\textcolor{black}{Average}}  
				& \textcolor{black}{57.61} & \textcolor{black}{63.29} & \textcolor{black}{52.94} & \textcolor{black}{56.60} & \textcolor{black}{69.05} & \textcolor{black}{32.99} & \textcolor{black}{29.26} & \textcolor{black}{48.47} & \textcolor{black}{75.76} & \textcolor{black}{72.38} & \textcolor{black}{\textbf{76.31}} & \textcolor{black}{\textbf{78.14}} \\
				\hline
				\multirow{6}{*}{\begin{tabular}[c]{@{}c@{}}Macro-F1\\ (\%)\end{tabular}} & C & \multirow{2}{*}{D} & 47.48 & 53.11 & 48.54 & \textcolor{black}{48.58} & 66.24 & 31.71 & 33.63 & 44.12 & \textbf{71.96} & 70.02 & 71.71 & \textbf{73.66} \\ 
				& A &  & 42.60 & 51.25 & 44.01 & \textcolor{black}{47.52} & 65.13 & 38.14 & 30.41 & 42.03 & 66.73 & 68.29 & \textbf{73.57} & \textbf{74.87} \\ \cline{2-15} 
				& D & \multirow{2}{*}{C} & 62.63 & 64.49 & 47.27 & \textcolor{black}{51.52} & 63.78 & 28.21 & 23.66 & 45.41 & 77.24 & 71.44 & \textbf{77.92} & \textbf{78.18} \\ 
				& A &  & 56.44 & 62.20 & 48.34 & \textcolor{black}{52.87} & 67.77 & 34.22 & 32.63 & 45.37 & 75.90 & 73.16 & \textbf{79.44} & \textbf{80.09} \\ \cline{2-15} 
				& D & \multirow{2}{*}{A} & 58.92 & 64.43 & 43.91 & \textcolor{black}{48.90} & 64.00 & 27.90 & 21.65 & 41.09 & \textbf{77.22} & 71.69 & 75.66 & \textbf{78.65} \\ 
				& C &  & 56.85 & 63.10 & 46.91 & \textcolor{black}{50.84} & 67.60 & 35.07 & 33.11 & 42.83 & \textbf{77.44} & 73.13 & 74.70 & \textbf{76.90} \\ \cline{2-15} & \multicolumn{2}{c}{\textcolor{black}{Average}}  
				& \textcolor{black}{54.15} & \textcolor{black}{59.76} & \textcolor{black}{46.50} & \textcolor{black}{50.04} & \textcolor{black}{65.75} & \textcolor{black}{32.54} & \textcolor{black}{29.18} & \textcolor{black}{43.48} & \textcolor{black}{74.42} & \textcolor{black}{71.29} & \textcolor{black}{\textbf{75.50}} & \textcolor{black}{\textbf{77.06}} \\\hline
		\end{tabular}
		\begin{tablenotes}\footnotesize
			\item[*] D: DBLPv7, C: Citationv1, A: ACMv9. The top 2 classification f1-scores are highlighted in bold for each task.
		\end{tablenotes}
	\end{threeparttable}}}
	\label{source_target_training_rate_10_5}
	\vspace{-0.5em}
\end{table*}

\subsection{Performance Comparison (RQ1)}\label{RQ1}

 The training rate of a network is defined as $R_l=\frac{|V^{l}|}{|V|}$, where $V^l$ represents the set of labeled nodes in the network. Different settings of $R_l$ are constructed by randomly sampling $V^l$ from $V$ while ensuring nodes in $V^l$ covering all labels. In this section, we conduct multi-label classification on three datasets with six transfer learning tasks. We consider two settings, an unsupervised setting with only the source network partially labeled, and a semi-supervised setting where both the source and target networks are partially labeled.
 
 \subsubsection{Unsupervised Setting: Partially Labeled Source Network and Completely Unlabeled Target Network}\label{unsupervised}
 In the unsupervised setting, we conduct experiments with the source training rate as 10\% while the target network is completely unlabeled. The experimental results are shown in Table~\ref{source_training_rate_10}. It can be easily observed that our proposed method AdaGCN outperforms all the baselines in five out of six tasks, and has comparable results with the best baseline CDNE on the sixth task Citationv1$\rightarrow$ACMv9. It demonstrates the effectiveness of our proposed AdaGCN model for cross-network node classification. Specifically, there is a 4.41\% relative performance improvement in Micro-F1 score and a 5.81\% in Macro-F1 score over the best baseline CDNE on average across all transfer tasks. AdaIGCN can further improve AdaGCN, and outperforms all the baselines consistently in all learning tasks.
 
 GCN and GraphSAGE have comparable performance. The proposed AdaGCN method adapts GCN for cross-network learning by combining it with domain adaptation technique. It achieves significant 13.54\% and 19.03\% relative gains in Micro-F1 and Macro-F1 scores respectively over GCN, which suggests that the adversarial domain adaptation component can effectively mitigate the distribution divergence of two domains and enables a successful knowledge transfer.
 The proposed AdaIGCN model further achieves significant 4.83\% and 4.28\% relative improvements in Micro-F1 and Macro-F1 scores respectively over AdaGCN on average, which shows that IGCN can learn better node representations to facilitate knowledge transfer.
 
 We noticed that both DeepWalk and node2vec have poor performance in all transfer learning tasks as shown in Table~\ref{source_training_rate_10}. The reason is that node representations used for multi-label classification are trained independently for the source and target networks since no connections between them exist. This makes the learned representations incomparable across networks, and thus the learned classifier based on source labeled data can not generalize to the target domain. Similar observations have also been made in~\cite{MLG-17-HeimannKoutra}. Therefore, single network embedding methods with only network topology as input are not directly suitable for multi-network learning. ANRL, as an attributed network embedding method, has much better performance compared with DeepWalk and node2vec, which benefits from the shared node attributes between the source and target networks. However, it is inferior to GCN by a large margin, not to mention the proposed AdaGCN method. The reasons lie in two aspects: firstly, ANRL is an unsupervised embedding method, so node classification can only be conducted after node representations have been learned, while GCN can perform semi-supervised learning in an end-to-end manner; secondly, ANRL suffers from the distribution shift between the source and target domains, while AdaGCN addresses this issue by introducing an adversarial domain adaptation component. \textcolor{black}{Another attributed network embedding method LANE also outperforms ANRL, since it incorporates label information in the embedding learning process.}
 
 Both DNNs and WDGRL cannot leverage network topology information. It can be observed that the performances of DNNs and WDGRL are poor, although more available labeled nodes can help improve their performances. Besides, we noticed that WDGRL performs worse than DNNs in some tasks, which means that the domain adaptation component of WDGRL results in negative transfer. The reason might be that the distribution divergence between node attributes of two domains are too large for the adversarial domain adaptation method to work. Overall, it suggests that existing domain adaptation methods can not handle cross-network node classification problem due to their inability in leveraging network structure information. In contrast, our proposed AdaGCN method jointly models network structures and node attributes with graph convolution. The Laplacian smoothing on node features with graph convolution in the representation learner enables an easy knowledge transfer across networks.
 
 NetTr and CDNE are two transfer learning methods for cross-network node classification. Our methods outperform NetTr by a large margin. Specifically, AdaGCN achieves remarkable 57.28\% and 76.46\% relative improvements over NetTr in Micro-F1 and Macro-F1 scores, respectively. One important reason is that NetTr learns transferable representations based on network topology only. Our proposed methods also produce a significant improvement over CDNE on average as mentioned before. Particularly, the relative performance gain of AdaGCN over CDNE reaches the desirable 12.47\% and 10.14\% in Micro-F1 and Macro-F1 scores respectively on ACMv9$\rightarrow$DBLPv7. The advantages of our methods over CDNE can be summarized into two aspects: firstly, graph convolution enables a natural combination of node attributes and network structures for representation learning, while CDNE only leverages network structures to extract features; secondly, the adversarial domain adaptation method is shown to be more effective compared with MMD in the literature~\cite{AAAI-ShenQZY18}.
 
 \subsubsection{Semi-Supervised Setting: Partially Labeled Source and Target Networks}
 
 In the semi-supervised setting, both of the source and target networks are partially labeled with 10\% and 5\% training rates, respectively. The results are shown in Table~\ref{source_target_training_rate_10_5}.
 
 Due to the additional available labeled data in the target network, all models achieve better classification performance compared with the unsupervised setting as shown in Tables~\ref{source_training_rate_10} and~\ref{source_target_training_rate_10_5}. There are many similar findings in both the unsupervised and semi-supervised scenarios, and we only highlight some new insights. Firstly, both DeepWalk and node2vec perform significantly better even though only 5\% additional labeled nodes in the target network are available. It shows the effectiveness of the learned node embeddings in the target network. Both GraphSAGE and GCN have better results compared with DeepWalk and node2vec because of the proper utilization of both node attributes and network topology in learning tasks and a certain level of knowledge transfer due to the shared weights in the representation learner. AdaGCN consistently outperforms GCN across all learning tasks by a large margin, which can be attributed to the successful knowledge transfer from the source to the target network thanks to the domain adaptation component. Similarly, AdaIGCN further improves over AdaGCN with 2.40\% and 2.04\% relative gains in Micro-F1 and Macro-F1 scores respectively because of the improved GCN layer for alleviating overfitting. It also produces 3.14\% and 3.55\% relative improvements in Micro-F1 and Macro-F1 scores respectively over the best baseline CDNE. 
 
 Overall, the empirical results demonstrate that our proposed methods achieve state-of-the-art cross-network node classification performance in both the unsupervised and semi-supervised settings. \textcolor{black}{To make our results more convincing, we also consider combining the source and target networks into a single network by randomly adding $\eta \in \{0, 10^{-5}\%, 10^{-3}\% \}$ cross-network edges and doing training and inference on the combined network with GCN. The  results reported in Table~\ref{gcn_ablation} show that even a very small number of artificial edges can lead to performance degradation, and our AdaGCN has a clear advantage.}    

\begin{table}[]
	\renewcommand{\arraystretch}{1.4}
	\textcolor{black}{\caption{
	Comparison with GCN on a combined network.
	}}
	\centering
	\textcolor{black}{\scalebox{0.85}{
	\centerline{
	\begin{threeparttable}
\begin{tabular}{cccccccc}
\hline
Settings           & Methods & C$\rightarrow$D            & A$\rightarrow$D            & D$\rightarrow$C            & A$\rightarrow$C            & D$\rightarrow$A            & C$\rightarrow$A            \\
\hline
\multirow{4}{*}{U} 
                  & GCN ($10^{-3}\%$)  & 64.93          & 61.78          & 56.08          & 67.67          & 52.96          & 58.12          \\
                  & GCN ($10^{-5}\%$)  & 67.40           & 65.20           & 59.68          & 70.52          & 55.22          & 62.65          \\
                  & GCN (0)  & 67.00             & 65.92          & 63.47          & 69.02          & 53.32          & 64.33          \\
                  & AdaGCN  & \textbf{70.89} & \textbf{69.32} & \textbf{77.77} & \textbf{78.83} & \textbf{67.92} & \textbf{68.30}  \\
                  \hline
\multirow{4}{*}{S}
                  & GCN ($10^{-3}\%$)  & 70.87          & 67.38          & 72.02          & 75.46          & 70.79          & 70.50           \\
                  & GCN ($10^{-5}\%$)  & 71.84          & 69.03          & 73.11          & 75.97          & 71.70           & 71.53          \\
                  & GCN (0)  & 71.25          & 70.29          & 73.09          & 75.28          & 71.51          & 72.84          \\
                  & AdaGCN  & \textbf{71.90}  & \textbf{75.18} & \textbf{79.66} & \textbf{81.45} & \textbf{75.15} & \textbf{74.51} \\
                  \hline
\end{tabular}
		\begin{tablenotes}\footnotesize
			\item[*] U: Unsupervised, S: Semi-supervised, D: DBLPv7, C: Citationv1, A: ACMv9. The best classification score in Micro-F1 ($\%$) is highlighted in bold for each task.
		\end{tablenotes}
	\end{threeparttable}}}}
	\label{gcn_ablation}
	\vspace{-0.5em}
\end{table}


 \subsection{Effect of Training Rate (RQ2)}\label{RQ2}

\begin{figure*}[t]
	\centering
	\includegraphics[width=2.0\columnwidth]{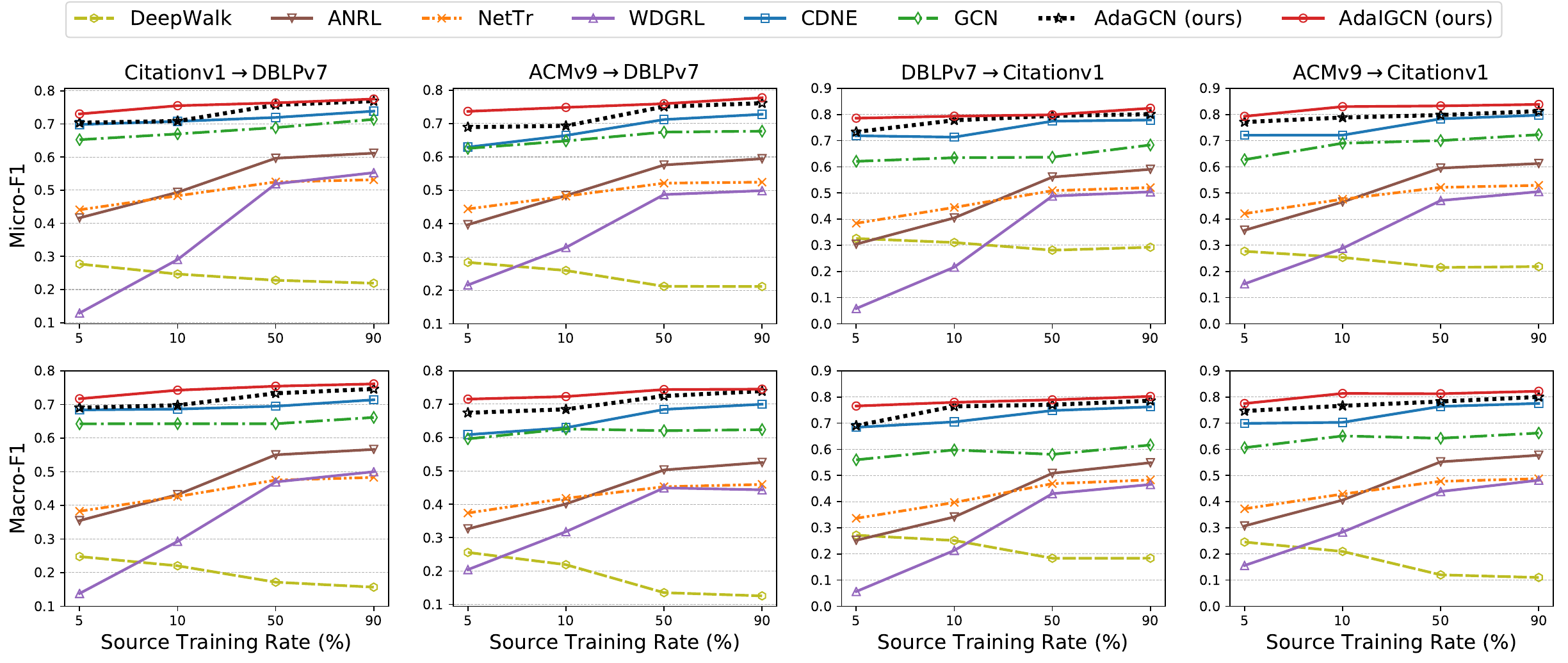}
	\caption{Multi-label classification with varying source training rates.}
	\label{souce_training_rate}
	\vspace{-0.5em}
\end{figure*}

 In this section, we study the effect of training rate $R_l$ of the source and target networks on model performance.
 
 \subsubsection{Effect of Source Training Rate}
 We conduct experiments with the training rate of source network ranging from 5\% to 90\% while the target network is completely unlabeled. The experimental results are displayed in Figure~\ref{souce_training_rate}. Note that only some of the baselines are selected for comparison to ensure clear presentations, and only the results on tasks with DBLPv7 and Citationv1 as targets are presented here to avioid repetition. We have the following observations:
  \begin{itemize}[leftmargin=0.3cm]
      \item Our proposed methods, including AdaGCN and AdaIGCN, consistently outperform all the baselines on these four tasks for all training rates, which demonstrates their effectiveness for knowledge transfer across networks. AdaIGCN performs better than AdaGCN, especially when the source training rate is low. It validates that the utilization of IGCN layer can help alleviate the overfitting issue and facilitate knowledge transfer.
      \item For almost all baselines except DeepWalk, the performance first improves, and then becomes stable with the increase of source training rate. For our proposed AdaIGCN, it shows remarkably good performance even with only 5\% labeled nodes in the source network, which suggests its high label efficiency.
      \item We noticed that the performance of DeepWalk decreases as the source training rate increases. It actually further confirms our finding that single network embedding methods based on topology only are not applicable for cross-network learning due to the incomparable node representations for two networks. Similar results can also be observed for node2vec which are not shown here.
  \end{itemize}

\begin{figure}[t]
	\centering
	\includegraphics[width=0.9\columnwidth]{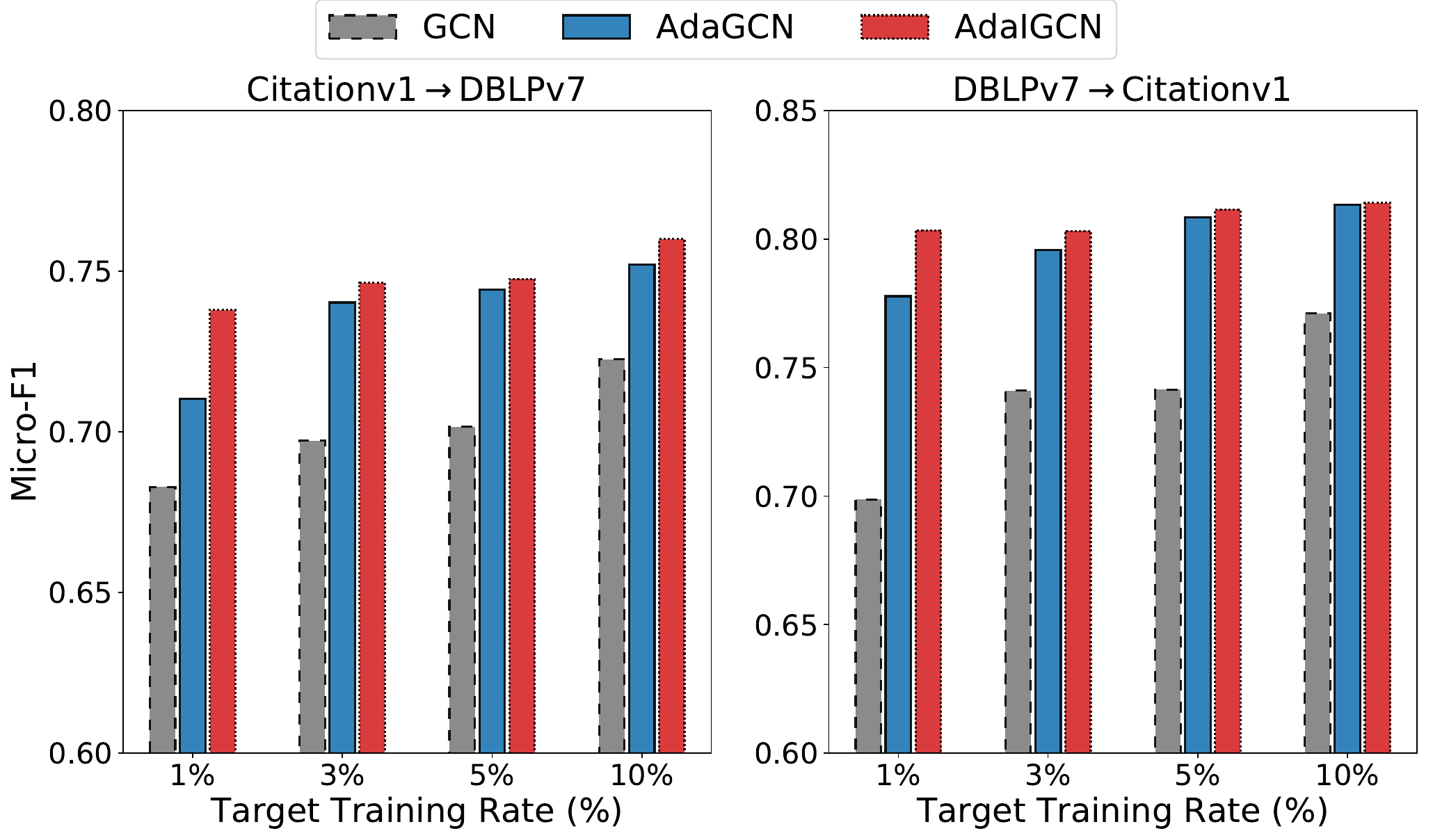}
	\caption{Multi-label classification with varying target training rates.}
	\label{fig:target_training_rate}
	\vspace{-0.5em}
\end{figure}

 \subsubsection{Effect of Target Training Rate}
 We investigate the effect of target training rate by varying it from 1\% to 10\% while fixing source training rate as 10\%. We only show the Micro-F1 scores in Figure~\ref{fig:target_training_rate} on learning tasks Citationv1$\rightarrow$DBLPv7 and DBLPv7$\rightarrow$Citationv1 for succinct presentation. We have the following observations. Firstly, AdaGCN significantly and consistently outperforms GCN on both learning tasks for all target training rates, which means that the adversarial domain adaptation component can successfully mitigate the distribution discrepancy between two domains and help knowledge transfer across networks. Specifically, AdaGCN exhibits an impressive 5.08\% relative improvement over GCN on average. Secondly, AdaIGCN further achieves improvements upon AdaGCN consistently, and the gap is more significant with low target training rate. In particular, it produces a 3.90\% relative gain over AdaGCN on Citationv1$\rightarrow$DBLPv7 when the target training rate is 1\%. It proves that the improved GCN layer can make a good balance between the strength of Laplacian smoothing and model complexity. Overall, it demonstrates the effectiveness of our proposed methods for \textcolor{black}{graph} transfer learning in the semi-supervised setting.

 \subsection{Effect of Distribution Discrepancy (RQ3)}
 
\begin{figure}[t]
	\centering
	\includegraphics[width=0.9\columnwidth]{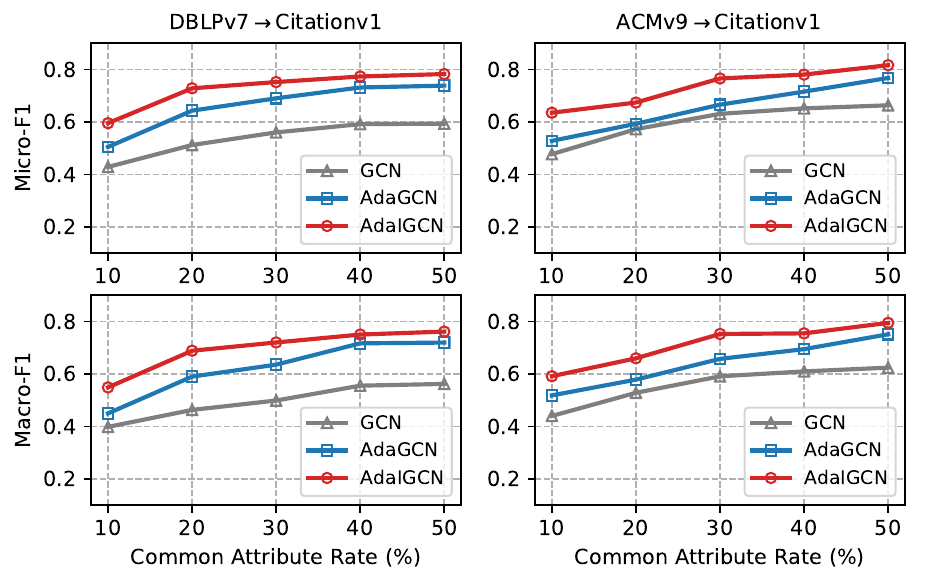}
	\caption{Multi-label classification on Citationv1 with varying common attribute rates of the source and target networks.}
	\label{common_attrb_rate}
	\vspace{-0.5em}
\end{figure}

 In this section, we explore the effect of distribution discrepancy between the source and target networks on domain adaptation. \textcolor{black}{We define common attribute rate between the source and target networks as $R_a=\frac{|\mathcal{X}^s\cap \mathcal{X}^t|}{|\mathcal{X}^s\cup \mathcal{X}^t|}$. When calculating $R_a$, the number of union attributes is fixed to be the one provided in Table~\ref{tab-dataset}, that is, 6775. In the experiments, we vary $R_a$ by randomly deleting some of the common attributes of two networks. Lower $R_a$ means larger distribution discrepancy. We conduct multi-label classification under unsupervised setting, with the source training rate set as 10\%. GCN, AdaGCN, and AdaIGCN are evaluated in transfer tasks where Citationv1 serves as the target network. In transfer tasks, “DBLPv7$\rightarrow$Citationv1” and “ACMv9$\rightarrow$Citationv1”, the initial common attribute rates are 55.84\% and 63.25\%, respectively.}
 
 Figure~\ref{common_attrb_rate} displays the experimental results when $R_a$ ranges from 10\% to 50\%. Both AdaGCN and AdaIGCN consistently outperform GCN across all common attribute rates for both transfer tasks. More specifically, AdaGCN achieves 22.88\% and 24.93\% relative gains on Micro-F1 and Macro-F1 scores respectively over GCN for DBLPv7$\rightarrow$Citationv1, and 9.02\% and 14.56\% for ACMv9$\rightarrow$Citationv1. It demonstrates that the adversarial domain adaptation component contributes to the classification performance even when the source and target networks only share a very small proportion of attributes. Besides, AdaIGCN performs better than AdaGCN consistently, which further confirms that the IGCN layer can learn better node representations for domain adaptation. In summary, the proposed methods are very robust and can work well with large distribution shifts between the source and target networks, which enables their applications for a wide range of real-world problems.

\subsection{Effect of Graph Convolution (RQ4)}

\begin{figure} \centering
	\subfigure[Parameter $n_I$.] { \label{fig:smoothing_steps}
		\includegraphics[width=0.43\columnwidth]{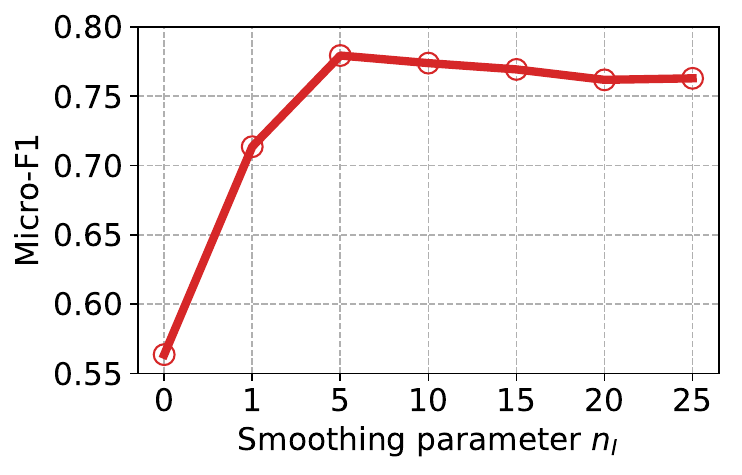}
	}
	\hspace{0.1in}
	\subfigure[Coefficient $\lambda$.] { \label{fig:da_coeff}
		\includegraphics[width=0.43\columnwidth]{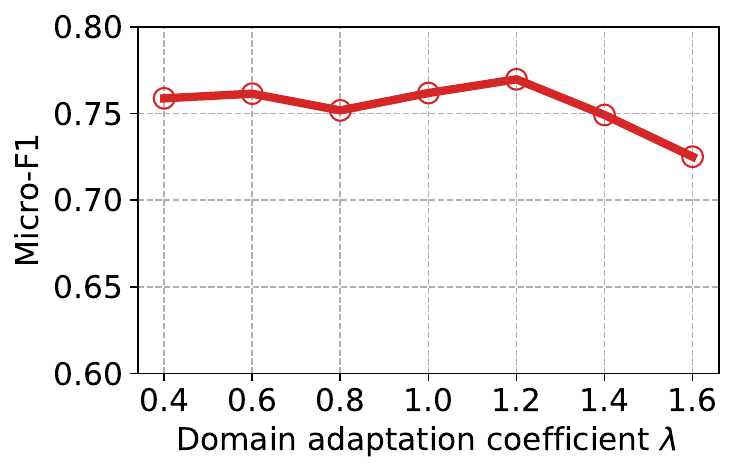}
	}
	\hspace{0.1in}
	\subfigure[Coefficient $\gamma$.] { \label{fig:gp_param}
		\includegraphics[width=0.43\columnwidth]{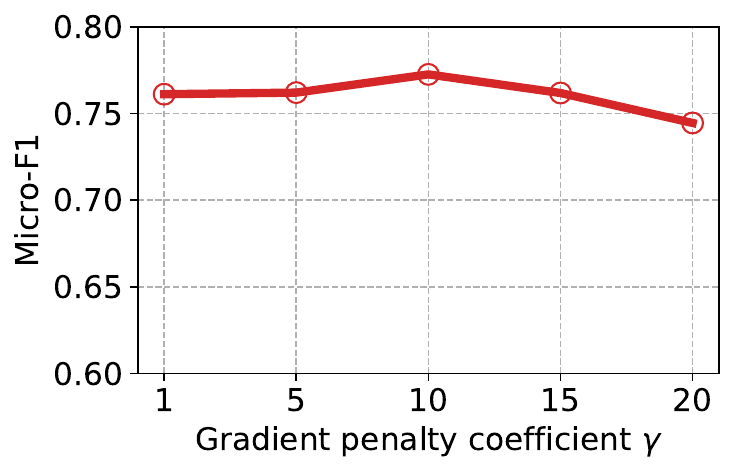}
	}
	\hspace{0.1in}
	\subfigure[Training step $n_d$.] { \label{fig:d_training_step}
		\includegraphics[width=0.43\columnwidth]{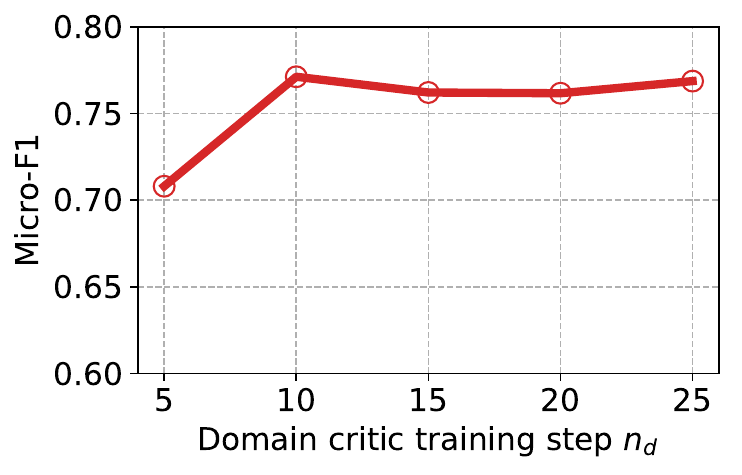}
	}
	\caption{Impact of hyper-parameters.}
	\label{fig:model-sensitivity}
	\vspace{-0.5em}
\end{figure}

 In this section, we vary the smoothing parameter $n_I$ of the IGCN layer in AdaIGCN from 1 to 25 to study the effect of graph convolution on domain adaptation. Note that AdaIGCN can be reduced to WDGRL when $n_I=0$, i.e., no smoothing on node features. The experiments are conducted in the unsupervised setting with source training rate as 10\%. We present the experimental results on DBLPv7$\rightarrow$Citationv1 in Figure~\ref{fig:smoothing_steps}. We can observe that graph convolution on node features brings extraordinary improvements to node classification performance on the target network, since there is a remarkable 26.62\% relative improvement when increasing $n_I$ from 0 to 1. When varying $n_I$ from 1 to 25, the classification accuracy first increases and then slightly drops. It shows that appropriate setting of $n_I$ can help further facilitate knowledge transfer, but too large $n_I$ can result in over-smoothing of node features and thus harm the transfer performance. Specifically, features of neighborhood nodes become similar with Laplacian smoothing in the graph convolutional layer, and a large smoothing parameter can make them converge to very similar value and blur the class boundaries. On the whole, graph convolution plays a crucial role for the successful knowledge transfer across networks in our proposed framework.

\subsection{Parameter Sensitivity (RQ5)}

 In this section, we perform sensitivity analysis of AdaGCN on domain adaptation coefficient $\lambda$, gradient penalty coefficient $\gamma$, and domain critic training step $n_d$. The experiments are conducted in the unsupervised setting with source training rate as 10\%. It is expected to shed some lights on how to configure these hyper-parameters. Here we only present the Micro-F1 score for Citationv1$\rightarrow$DBLPv7 to avoid repetition, and similar tendency can be observed in other tasks. Note that when studying one hyper-parameter, we fix all others with default settings mentioned in Section~\ref{parameter_settings}.
  
 $\lambda$ is a coefficient for balancing the semi-supervised loss and domain adaptation loss. We can find that the performance slightly improves with the increase of $\lambda$ from 0.4 to 1.2, and then drops quickly afterwards as shown in Figure~\ref{fig:da_coeff}. It suggests that it is important to maintain the balance between the two parts so as to learn both class discriminative and domain invariant representations. $\gamma$ is a hyper-parameter for controlling the weight of gradient penalty when training the discriminator of the adversarial domain adaptation component. From Figure~\ref{fig:gp_param}, it can be observed that the best result is obtained when $\gamma$ is set to 10, and smaller or larger configurations might result in performance degradation, which is consistent with the finding in~\cite{NIPS-GulrajaniAADC17}, and thus 10 would be a recommended setting. Theoretically, the domain critic network $f_d(\cdot)$ should be trained to optimality by optimizing its own parameters while fixing those of other components, and thus the training step $n_d$ should be set to a large enough number for this purpose. From Figure~\ref{fig:d_training_step}, it can be noticed that the Micro-F1 score shows apparent increase when increasing $n_d$ from 5 to 10, and then becomes stable, which is consistent with our theoretic analysis.

\subsection{Visualization of Node Representations}
 \begin{figure}[t] \centering
	\subfigure[GCN.] { \label{fig:gcn}
		\includegraphics[width=0.4\columnwidth]{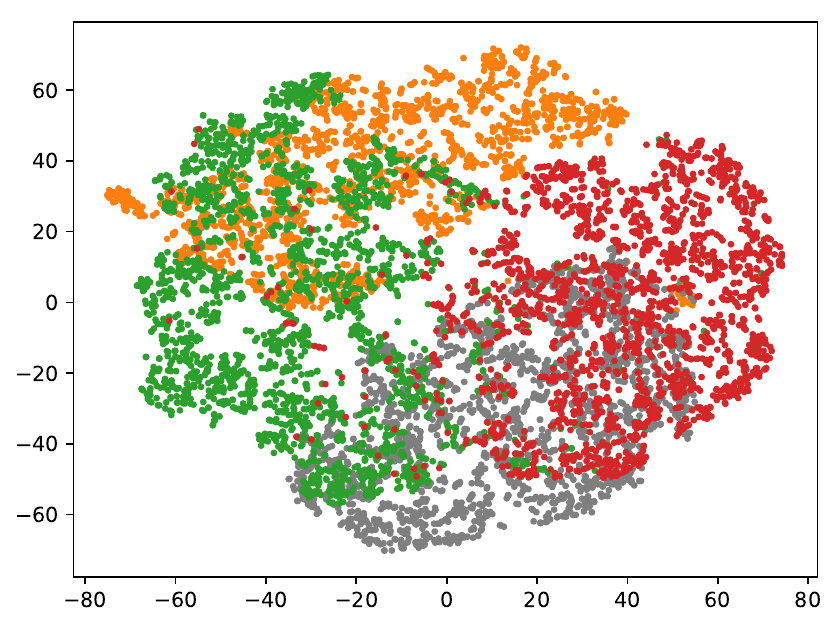}
	}
	\hspace{-0.08in}
	\subfigure[IGCN.] { \label{fig:igcn}
		\includegraphics[width=0.4\columnwidth]{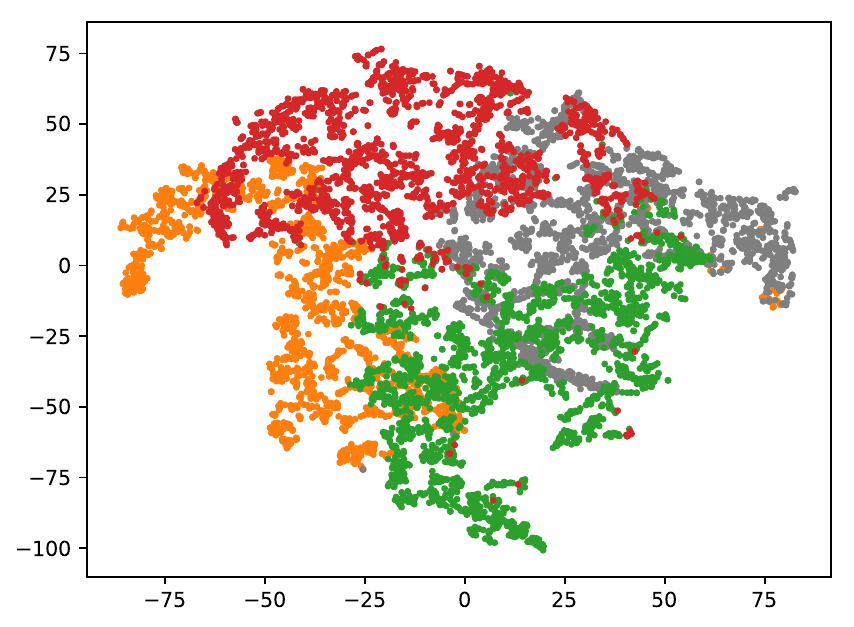}
	}
	\hspace{-0.08in}
	\subfigure[AdaGCN.] { \label{fig:adagcn}
		\includegraphics[width=0.4\columnwidth]{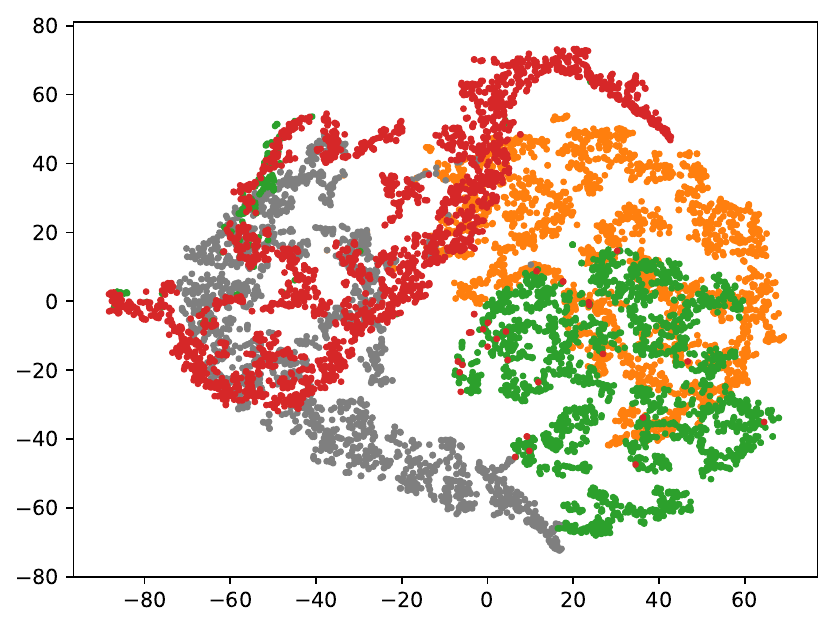}
	}
	\hspace{-0.08in}
	\subfigure[AdaIGCN.] { \label{fig:adaigcn}
		\includegraphics[width=0.4\columnwidth]{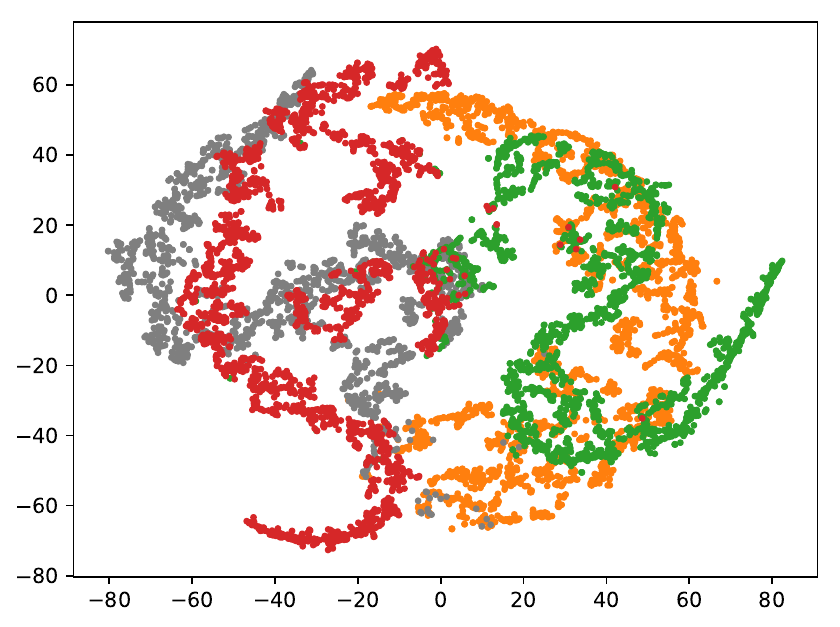}
	}
	\caption{Visualization of the learned node representations from ACMv9$\rightarrow$Citationv1. Each point represents one paper. Gray and orange points are from the source network, and red and green points are from the target network. Gray and red: ``Databases''. Orange and green: ``Computer Vision''. These plots are best viewed in color.}
	\label{acmv9_citationv1_visualization}
	\vspace{-0.5em}
 \end{figure}
 
 Figure~\ref{acmv9_citationv1_visualization} visualizes the node representations generated by GCN, IGCN, AdaGCN and AdaIGCN in the unsupervised setting for ACMv9$\rightarrow$Citationv1 using t-SNE~\cite{JMLR-08-Maaten} where the source network is fully labeled. We only visualize nodes from ``Databases'' and ``Computer Vision'' for clear presentation. The gray and orange points represent papers of ``Databases'' and ``Computer Vision'' respectively from ACMv9, while red and green points represent papers of ``Databases'' and ``Computer Vision'' from Citationv1.
 
 On one hand, the domain adaptation component helps mitigate domain divergence and benefits knowledge transfer. Specifically, from Figures~\ref{fig:gcn} and~\ref{fig:igcn}, it can be observed that both the GCN and IGCN models suffer from distribution shift between different networks, since nodes from different categories, e.g., green and gray points, are mixed together. In contrast, from Figures~\ref{fig:adagcn} and~\ref{fig:adaigcn}, we can find that gray and red points are clustered together, while orange and green nodes are clustered together. It demonstrates that the adversarial domain adaptation successfully mitigates the distribution divergence between the source and target networks, since papers from the same categories of both domains are well clustered together. Besides, the boundary between these two clusters are quite clear, which means that the learned node representations are discriminative. On the other hand, the IGCN layer also brings two significant advantages. 
 Firstly, the IGCN layer allows adjusting the smoothing strength on node features without increasing model complexity, and an appropriate smoothing of node features helps to learn more compact node representations within the same category as shown in Figures~\ref{fig:igcn} and~\ref{fig:adaigcn}, thus contributes to the classification task. Furthermore, it makes the domain adaptation process easier, which is confirmed by the visualization results that AdaIGCN aligns the source and target node representations better than AdaGCN as shown in Figures~\ref{fig:adagcn} and~\ref{fig:adaigcn}.
 

\section{Conclusion} \label{conclusion}
 In this paper, we successfully address the cross-network node classification problem by proposing a novel \textcolor{black}{graph} transfer learning framework AdaGCN, which leverages the techniques of adversarial domain adaptation and graph convolution. It can learn both class discriminative and network invariant node representations with the help of a semi-supervised learning (SSL) component and an adversarial domain adaptation (ADA) component. The SSL component is capable of learning a well-generalized node classifier with graph convolutional layers for representation learning, while the ADA component ensures successful knowledge transfer from the source network to the target network through adversarial learning. Together they enable AdaGCN to work well in real-world attributed networks under a realistic setting.

 \textcolor{black}{The research of transfer learning on networked data is still in an early stage, and much more effort is needed.} This paper serves as a step further in this direction. Future work will include investigating knowledge transfer from multiple source networks to a target network and exploring conditional adversarial domain adaptation for better alignment of multimodal data distribution.
	
	\ifCLASSOPTIONcompsoc
	\section*{Acknowledgments}
	\else
	\section*{Acknowledgment}
	\fi

	This research was partially supported by National Natural Science Foundation of China (No. 62102124), HK ITF UIM/363 and the grants 1-ZVJJ and G-YBXV funded by the Hong Kong Polytechnic University.

	\ifCLASSOPTIONcaptionsoff
	\newpage
	\fi

	
	
	\bibliographystyle{IEEEtran}
	\bibliography{Embeddings}
	
	\begin{IEEEbiography}[{\includegraphics[width=1in,height=1.25in,clip,keepaspectratio]{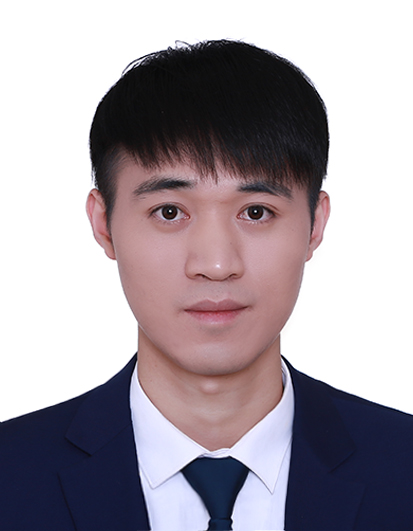}}]{Quanyu Dai}
		is currently a researcher at Huawei Noah's Ark Lab. He received the B.Eng. degree in information engineering from Shanghai Jiao Tong University, in 2015, and the Ph.D. degree at the Department of Computing, The Hong Kong Polytechnic University, in 2020. His research interests include machine learning, data mining and recommender systems. He has publications appeared in the top-tier journals and conferences, such as TKDE, TNNLS, IJCAI, AAAI, WWW and KDD.
	\end{IEEEbiography}
	
	\vspace{-0.3in}
	\begin{IEEEbiography}[{\includegraphics[width=1in,height=1.25in,clip,keepaspectratio]{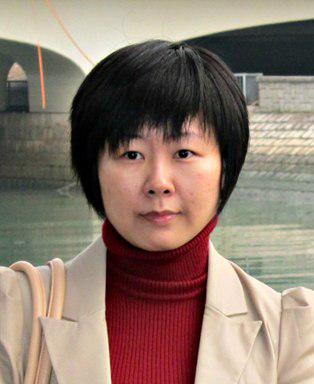}}]{Xiao-Ming Wu} is currently an assistant professor at the Department of Computing, The Hong Kong Polytechnic University. She obtained her PhD degree from the Department of Electrical Engineering, Columbia University in 2016. Prior to that, she obtained her MPhil degree from the Chinese University of Hong Kong and her bachelor’s and master’s degrees from Peking University. Her research interests include machine learning and applications of artificial intelligence in computer vision, natural language processing, and search and recommendation. 
	\end{IEEEbiography}
	
	\vspace{-0.3in}
	
	\begin{IEEEbiography}[{\includegraphics[width=1in,height=1.25in,clip,keepaspectratio]{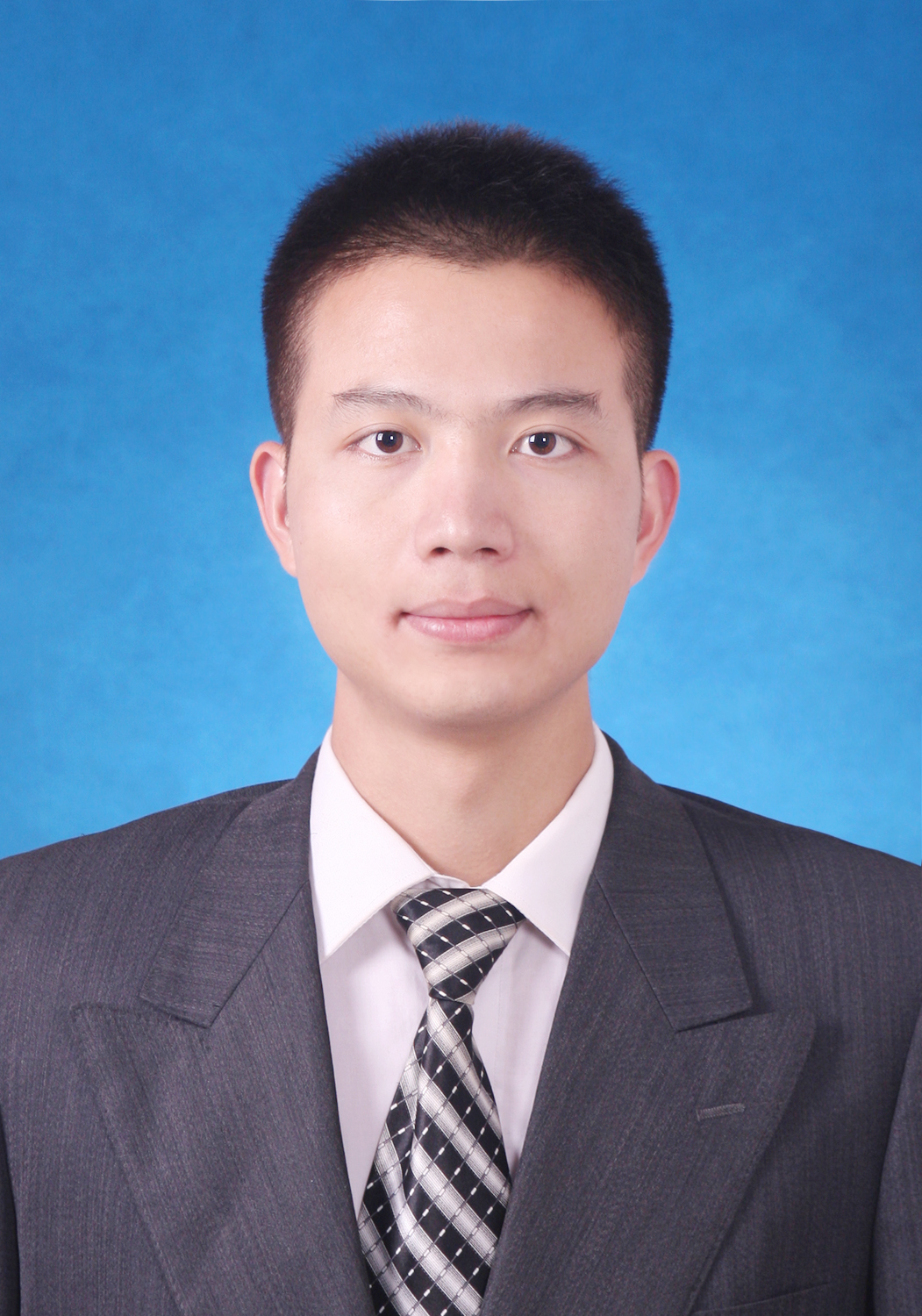}}]{Jiaren Xiao} received the B.Eng. degree in mechanical engineering from Xi'an Jiaotong University, Xi'an, China, in 2015, and the M.Eng. degree in mechanical engineering from Shanghai Jiao Tong University, Shanghai, China, in 2018. He is now a Ph.D. candidate at the Department of Mechanical Engineering, The University of Hong Kong, Hong Kong, China. His research interests include graph representation learning and transfer learning.	
   \end{IEEEbiography}
	
	\vspace{-0.3in}
	
	\begin{IEEEbiography}[{\includegraphics[width=1in,height=1.25in,clip,keepaspectratio]{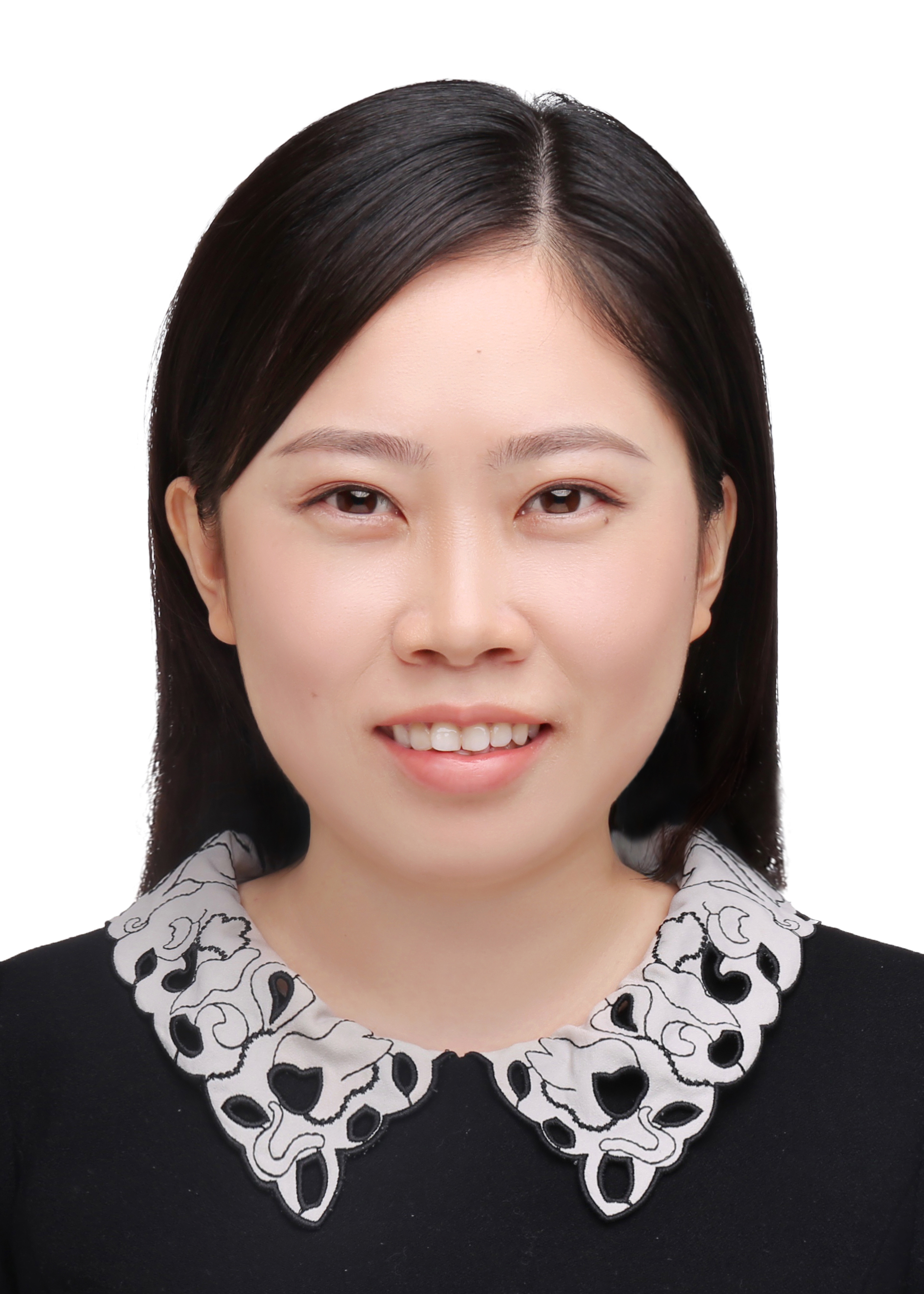}}]{Xiao Shen}
		received the double B.Sc. degrees from Beijing University of Posts and Telecommunications and Queen Mary University of London in 2012, the M.Phil. degree from University of Cambridge in 2013, and the Ph.D. degree from Department of Computing, Hong Kong Polytechnic University, in 2019. She was a Postdoc Fellow at Hong Kong Polytechnic University. She is now an Associate Professor with Hainan University, China. She received the Hong Kong PhD Fellowship. Her research interests include feature representation learning, deep learning, transfer learning and data mining in complex networks.
	\end{IEEEbiography}

	\vspace{-0.3in}

	\begin{IEEEbiography}[{\includegraphics[width=1in,height=1.25in,clip,keepaspectratio]{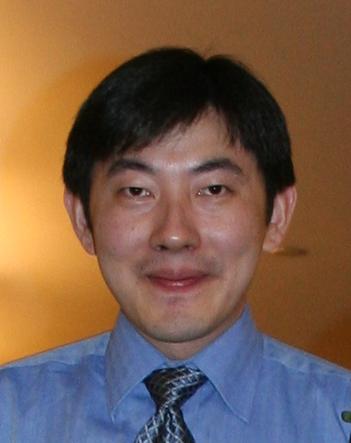}}]{Dan Wang}
		is currently an associate professor at Department of Computing, The Hong Kong Polytechnic University. He is an expert in computer networking, and he is recently working in the inter-discipline domains of smart energy systems, industry 4.0. He publishes extensively in top networking conferences, such as ACM SIGCOMM, ACM SIGMETRICS, IEEE INFOCOM and in top inter-discipline conference, such as ACM e-Energy, ACM Buildsys. He won the Best Paper Award of ACM e-Energy 2018, the Best Paper Award of ACM Buildsys 2018. He will serve as the TPC co-Chair of ACM e-Energy 2020. Dan Wang received his B.Sc degree from Peking University, China, his M.Sc degree from Case Western Reserve University, Cleveland, Ohio, and his Ph.D. degree from Simon Fraser University, Canada, all in computer science.
	\end{IEEEbiography}

	
	

\end{document}